\pdfoutput=1
\documentclass[11pt]{article}

\PassOptionsToPackage{table}{xcolor}
\usepackage[final]{acl}

\usepackage{times}
\usepackage{latexsym}

\usepackage[T1]{fontenc}

\usepackage[utf8]{inputenc}

\usepackage{microtype}

\usepackage{inconsolata}

\usepackage{graphicx}

\usepackage{placeins} 
\usepackage{dblfloatfix}
\usepackage{hyperref}
\usepackage{url}
\usepackage{makecell}
\usepackage{amsmath}
\usepackage{amssymb}
\usepackage{wrapfig} 
\usepackage{multirow}
\usepackage{tabularray}
\UseTblrLibrary{booktabs}
\usepackage[normalem]{ulem} 
\usepackage{tikz}
\usepackage{arydshln}
\usepackage{tcolorbox}
\tcbuselibrary{breakable}
\usepackage{wrapfig}
\usepackage{cleveref}   
\usepackage{pifont} 
\usepackage{algorithm}
\usepackage{algpseudocode}
\usepackage{subcaption}  
\usepackage{caption}     
\usepackage{fvextra}

\title{Beyond Rubrics: Exploration-Guided Evaluation Skills for Reward Modeling}

\author{Xing Yue$^{1}$ \quad Linjuan Wu$^{1,2}$\thanks{\ \ Work done during internship at Xiaohongshu.} \quad Daoxin Zhang$^{2}$ \quad Yongliang Shen$^{1}$ \quad  Weiming Lu$^{1}$\thanks{\ \ Corresponding author.} \\
  $^1$Zhejiang University \quad $^2$Xiaohongshu~Inc.\\
  \texttt{\{yue\_xing, wulinjuan525, syl, luwm\}@zju.edu.cn} \\
  \texttt{tangxiaohui@xiaohongshu.com} \\
  }

\begin{document}
\maketitle
\begin{abstract}
Open-ended reward modeling requires judges that can follow subtle, domain-specific preferences when verifiable answers are unavailable. Existing rubric-based methods often address this by generating criteria online for each query, but the extra generation step can add inference overhead and produce rigid or misaligned guidance. We introduce \texttt{Eval-Skill}, an exploration-guided method that synthesizes reusable evaluation skills for reward modeling and reframes reward guidance as context evolution rather than parameter training or per-query rubric generation. Using only 100 cases per domain for skill evolution, \texttt{Eval-Skill} synthesizes reusable domain-level evaluation skills through two progressive stages, workflow generation followed by principle generation, with exploration and selection interleaved across both stages. Once generated, a skill is directly injected into the judge context. Across multiple RM benchmarks, \texttt{Eval-Skill} consistently improves diverse judge backbones; on RewardBench~2, it yields significant gains over vanilla judging for each main backbone (+13.44\% for Qwen3-8B, and 18.51\% for DeepSeek-V4-Flash). Further analyses of evolution-time scaling, generalizability, and transferability show that compact evaluation skills offer an efficient new paradigm for LLM-based evaluation. Code is available at \url{https://github.com/xing-stellus-yue/Eval-Skill}.
\end{abstract}

\section{Introduction}
\label{sec:introduction}
Reliable evaluation has become a central bottleneck for large language models (LLMs), especially in open-ended tasks where answers cannot be verified automatically. LLM-as-a-judge (LAAJ) methods are widely used as benchmark evaluators \citep{zheng2023judging}, reward models (RMs) for reinforcement learning and preference alignment \citep{ouyang2022training}, and verifiers for test-time scaling methods such as sequential self-reflexion \citep{shinn2023reflexion}. Yet strong judging requires more than a capable backbone: the judge must apply the right evaluation criteria, comparison procedure, and preference priorities for each task.

\begin{figure}[!t]
\setlength{\abovecaptionskip}{5pt}
\setlength{\belowcaptionskip}{-20pt}
    \centering
    \includegraphics[width=1.0\linewidth]{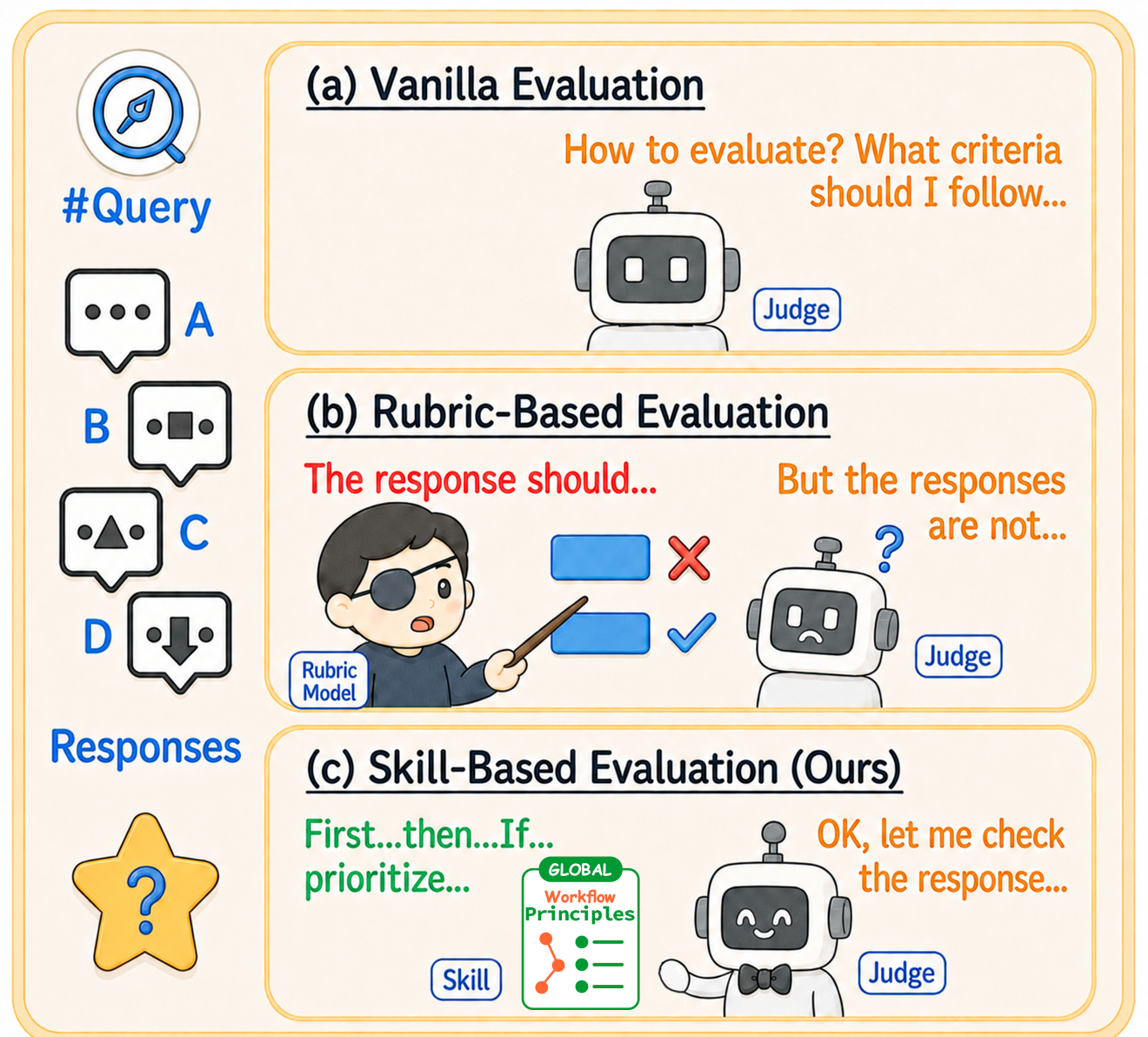}
    \caption{Comparison of (a) vanilla evaluation, (b) rubric-based evaluation, and (c) our skill-based evaluation.}
    \label{fig:teaser}
\end{figure}

Rubrics have been used in two related but distinct ways. In benchmark design, human-written rubrics are part of the evaluation specification and can improve evaluation in specialized benchmarks \citep{arora2025healthbench,sharma2025researchrubrics}. In contrast, online rubric-augmented reward modeling is an RM method: it generates a fresh rubric from each query at inference time and then asks a downstream judge to score or compare responses under that rubric \citep{liu2025openrubrics,xu2026alternating,dhole2026rubricrag}. Our critique targets this second paradigm, not benchmark-provided rubrics. Although online rubric generation is intuitive, it has two limitations for practical reward modeling. First, generating a new rubric for every query introduces an additional inference step. Second, because the rubric is usually generated from the query alone, it can miss response-dependent differentiators, over-specify surface-level requirements, or impose rigid criteria that mislead the judge. In our preliminary experiments, such generated rubrics can even make judges underperform vanilla one-step baselines on RewardBench \citep{lambert2025rewardbench} and RewardBench~2 \citep{malik2025rewardbench}. Motivated by the RIFT taxonomy \citep{qi2026rift}, we analyze these failures in \S~\ref{sec:r2s}.

We instead ask whether reward-model guidance can be evolved once and reused. Inspired by recent skill-based methods \citep{anthropic_skills,xia2026skillrl,wang2026skillx,li2026arise,ouyang2026skillos,wang2026effiskill}, we study \emph{skill-based reward modeling}. A skill is a domain-level context artifact generated offline from rollouts on a small evolving set and then directly injected into the judge prompt at test time. Unlike a case-specific rubric, a skill can encode not only criterion-like principles, but also an evaluation workflow: how to compare candidates, when to prioritize factuality or instruction following, how to handle ties, and how to apply conditional rules. This makes skills a more expressive and efficient form of guidance: once a skill is synthesized, no additional rubric model or per-query generation step is needed.

However, skill quality is critical. We find that a naive skill can already outperform online rubrics, but a poorly aligned workflow can also degrade performance. Moreover, simple iterative refinement, which is effective in some other skill-based settings \citep{zhang2026coevoskills}, does not reliably improve reward modeling. Preference-sensitive evaluation requires searching over competing judgment procedures rather than merely accumulating more instructions. We therefore argue that exploration and selection are essential for synthesizing strong RM skills.

We introduce \texttt{Eval-Skill}, an exploration-guided method for synthesizing reusable evaluation skills for reward modeling without parameter training. \texttt{Eval-Skill} constructs each skill progressively in two stages, workflow generation followed by principle generation, while interleaving exploration and selection across both stages. Since skill evolution is performed offline, this design separates skill-improvement compute from inference-time judging.

For each domain, we synthesize a skill from only 100 evolving cases and evaluate it on held-out test cases. Across RewardBench~2, RewardBench, and RM-Bench \citep{liu2024rmbench}, \texttt{Eval-Skill} improves diverse judge backbones and transfers across related domains and models. On RewardBench~2, it improves over vanilla one-step judging by 11.75, 13.44, and 18.51 percentage points for Qwen3-4B, Qwen3-8B, and DeepSeek-V4-Flash, respectively. Indeed, scaling the offline sampling and selection process can further improve performance without increasing inference-time cost (\S~\ref{ssec:sets}). We also find that generated skills transfer across judge backbones, allowing a skill evolved from one model's rollouts to benefit other models (\S~\ref{ssec:other_backbones}). Together, these results show that compact, reusable evaluation skills provide a high-performing and inference-efficient paradigm for reward modeling.

Our contributions can be summarized as follows:
\begin{itemize}
    \item We identify limitations of online rubric generation for reward modeling and recast evaluation guidance as reusable domain-level skills that can be directly injected into the judge context.
    \item We introduce \texttt{Eval-Skill}, a skill-based reward modeling method that automatically synthesizes high-quality evaluation skills through two-stage construction and exploration-selection.
    \item We show that \texttt{Eval-Skill} delivers 10+ percentage-point gains across main judge backbones, improves multiple RM benchmarks, and also benefits reward-guided best-of-$N$ inference. Our analyses of evolution-time scaling, transferability, and mixed-domain settings provide insights for skill-based evaluation methods and beyond.
\end{itemize}

\section{From Rubrics to Skills}
\label{sec:r2s}
\subsection{Task Formulation and Online Rubric-Based Reward Modeling}
\label{ssec:task_formulation_rubric_based_method}
Reward modeling can be pointwise, pairwise, or listwise; in this work, we focus on pairwise and listwise settings. Formally, given a query $q$ and a set of candidate responses $Y=\{y_1, y_2, \ldots, y_n\}$, a reward model $M_\text{reward}$ selects the best response $y_\text{r}$ from $Y$:
\[
y_\text{r} = M_\text{reward}(q,Y)
\]
For an online rubric-based RM method, reward modeling proceeds in two steps: a rubric model $M_\text{rubric}$ first generates a query-specific rubric, and then $M_\text{judge}$ selects $y_\text{r}$ under the guidance of that generated rubric:
\[
R=M_\text{rubric}(q), y_\text{r}=M_\text{judge}(q,Y,R)
\]
Ideally, the rubric $R$, composed of a set of criteria $c_i$, should help $M_\text{judge}$ make a better judgment.

\subsection{Limitation of Online Rubric Generation}
\label{ssec:rubric_based_methods}
Although online rubric generation is promising as an RM method, our experiments show that it can also substantially degrade LAAJ performance. As shown in Table~\ref{tab:rubric_failure}, directly adding generated rubrics often lowers accuracy on RewardBench~2 and RewardBench. For example, Qwen3-8B's accuracy on RewardBench~2 drops from 57.04\% to 50.63\% when it is guided by self-generated rubrics. Similarly, for Rubric-ARM-8B, a fine-tuned two-model system composed of a rubric model and a judge model, using the judge alone performs better than using the full rubric-guided system. 

\begin{table}[!h]
\setlength{\belowcaptionskip}{-12pt}
\centering
\footnotesize
\begin{tblr}{
width=\linewidth,
colspec={X[1.8,l] X[1.8,l] X[1,c] X[1,c]},
colsep=2pt,
rowsep=1pt,
rows={valign=m},
hline{5,7}={0.3pt,dashed},
hline{1,Z}={1pt,solid},
hline{2}={0.5pt,solid},
row{1}={font=\bfseries},
}
{Judge Model} & {Rubric Model} & RB~2 & RB \\
Qwen3-8B & w/o & \textbf{57.04} & \textbf{82.39} \\
Qwen3-8B & Qwen3-8B & 50.63 & 78.01 \\
Qwen3-8B & Rubric-ARM-8B & 51.57 & 78.60 \\
Rubric-ARM-8B & w/o & \textbf{62.01} & \textbf{85.46} \\
Rubric-ARM-8B & Rubric-ARM-8B & 60.79 & 79.37 \\
DS-V4-Flash & w/o & 68.27 & \textbf{89.66} \\
DS-V4-Flash & DS-V4-Flash & \textbf{68.69} & 88.00 \\
\end{tblr}

\caption{Comparison of the vanilla method (without rubric) and online rubric-based RM methods. DS denotes DeepSeek, and RB denotes RewardBench.}
\label{tab:rubric_failure}
\end{table}

Following the RIFT taxonomy for analyzing rubric failures \citep{qi2026rift}, we sample 200 failed rollouts from each RewardBench~2 domain, all produced by a Qwen3-8B judge with Qwen3-8B-generated rubrics, and use DeepSeek-V4-Flash to categorize their failure modes. The results in Figure~\ref{fig:rubric_failure_modes} show that ``Missing Criteria'' (34.2\%) and ``Misaligned or Rigid'' (24.1\%) are frequent failure modes. Through case analysis, we find that these failures are mainly caused by the response-agnostic design of the rubric generator: because the rubric is generated from the query alone, it can over-specify surface-level requirements, miss response-dependent differentiators, or impose rigid criteria that do not match the actual candidate responses. We tried many prompt variants, such as specifying that the LAAJ should treat the rubrics as ``merely an AI-generated reference''. However, these changes are not sufficient to prevent misleading guidance.

Beyond response agnosticism, we also find that the criterion-list format commonly used by online rubric generation is not sufficiently expressive for reward modeling. Such a generated rubric consists of a \emph{set} of criteria, which are usually applied individually and unconditionally. This format supports criterion-by-criterion scoring of candidate responses, but makes more complex LAAJ workflows difficult to express, such as conditional application, branching by query type, or head-to-head comparison. It also lacks accompanying components such as examples.

\subsection{A Naive Skill-Based Method}
\label{ssec:naive_skill_method}
Inspired by recent work on skills \citep{ni2026traceskill,zhang2026coevoskills}, we investigate skill-based methods for reward modeling. We frame a skill as a superset of a rubric: it can include criterion-like principles as well as a proper workflow\footnote{In fact, a skill could also contain a workflow specifying that the judge should first generate a rubric itself and then apply it to the candidate responses.} that guides the judge to perform reward modeling in a task-appropriate way. Whereas online generated rubrics are case-specific and must be regenerated for new cases, skills are domain-specific, generated offline, and maintained as one skill per domain:
\[
y_\text{r}=M_\text{reward}(q,Y,s)
\]
where $s$ denotes the domain-level skill. This design mitigates several limitations of online rubric-generation methods: (1) the increased latency caused by two-step generation for new cases; (2) the risk of generating misleading rubrics; and (3) the limited expressivity of the criterion-list rubric structure.
Although both rubrics and skills are forms of guidance injected into the judge prompt, a static domain-level skill, even a naive one, can significantly outperform seemingly more dynamic online rubrics, as shown in Table~\ref{tab:main_results}.

\section{Methodology}
\label{sec:method}

\begin{figure*}[!t]
\setlength{\abovecaptionskip}{5pt}
\setlength{\belowcaptionskip}{-12pt}
    \centering
    \includegraphics[width=\linewidth]{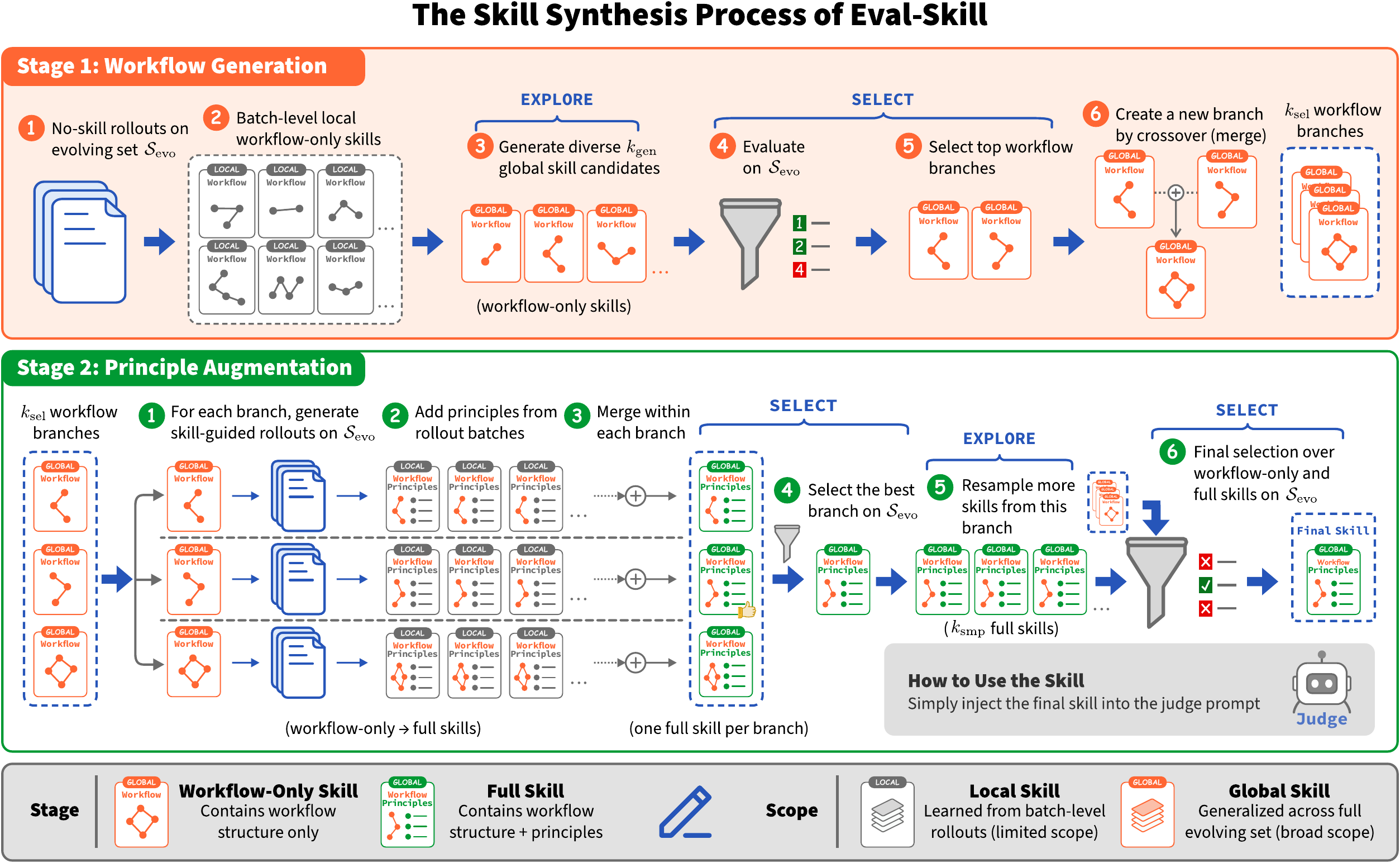}
    \caption{\texttt{Eval-Skill} synthesizes skills in two stages, workflow followed by principles, while interleaving exploration and selection.}
    \label{fig:fig_pipeline}
\end{figure*}

\subsection{Skill System Design}
\label{ssec:skill_system_design}

\paragraph{Skill Structure}
For compatibility, we follow the skill template introduced by Anthropic \citep{anthropic_skills}: a Markdown file containing YAML front matter and a body. Based on our analysis in \S~\ref{ssec:naive_skill_method}, the body of our skill mainly consists of two parts: (1) a high-level \texttt{Workflow}, which instructs the judge on how to evaluate, and (2) low-level \texttt{Principles}, which tell the judge what to do under specific conditions. We also include a one-sentence overview at the start of the body.

Decoupling \texttt{Workflow} and \texttt{Principles} enables different generation strategies, as described below.

\paragraph{One Skill Per Domain}
Some prior work maintains a skill library composed of multiple skills and retrieves skills at inference time. Because skills are often generated from a single rollout or a small batch of rollouts, such ``local skills'' can be specific and informative, but they can also misalign with test queries. Current retrieval methods based on embedding similarity may match skills from cases that are lexically similar but require different evaluation priorities, leading to ineffective or misleading guidance. We therefore follow Trace2Skill \citep{ni2026traceskill} by merging local skills and maintaining one global domain-level skill. This design avoids additional embedding or reranker models and empirically performs well.

\subsection{\texttt{Eval-Skill}}
Different workflows can encode different priorities or evaluation styles, especially for reward modeling, where an ideal workflow may be harder to infer from rollouts than in agent tasks \citep{ni2026traceskill}. It is therefore useful to explore multiple workflows and select the best one. At the same time, merging is suitable for consolidating a set of principles or rubrics, but not necessarily for workflows: two workflows with contrasting priorities or different first steps can be difficult to merge without distortion, and merging alone does not provide the exploration we need.

Inspired by genetic algorithms, we introduce \texttt{Eval-Skill}, an exploration-guided method that generates skills progressively in two stages, first workflows and then principles, while interleaving both stages with exploration and selection (see Figure~\ref{fig:fig_pipeline}). More details can be found in Appendix~\ref{apx:eval_skill_details}.

\paragraph{Workflow Generation}
We start by using the judge model $M_\text{judge}$ (without a skill) to generate rollouts on a small evolving set $\mathcal{S}_\text{evo}$. We divide these rollouts into batches and use them to extract batch-level local skills independently. Specifically, we use a strong model $M_\text{skill}$ for skill generation, conditioned on the query, the candidate responses, the rollout, the ground truth, and an LLM-generated answer explanation for each case. Skills generated in this stage are workflow-only. We generate local skills in two rounds. In the first round, we use rollouts from a stronger model, without skill guidance, as supervision. In the second round, we use rollouts from $M_\text{judge}$ guided by the previous-round skill to further refine the skill.

Instead of simply merging these skills into a global one, we use their workflows as references and instruct $M_\text{skill}$ to generate $k_\text{gen}$ \emph{diverse} new workflows, encouraging exploration of different priorities and processes while solving the same task\footnote{We discuss the methodology of consolidation in \S~\ref{ssec:ablation}.}. After wrapping each generated workflow into a workflow-only skill, we select the top $k_\text{sel}^{\prime}$ skills based on their performance on the full $\mathcal{S}_\text{evo}$. Similar to the crossover process in genetic algorithms, we create one additional skill by merging the $k_\text{sel}^{\prime}$ selected skills. This yields $k_\text{sel}=k_\text{sel}^{\prime}+1$ workflow-only skills $S_\text{W}$ in total, each forming an independent candidate branch for later selection.

\paragraph{Principle Augmentation}
For each workflow branch $s_\text{W} \in S_\text{W}$, we again generate rollouts on $\mathcal{S}_\text{evo}$ using $M_\text{judge}$ equipped with $s_\text{W}$, and divide the rollouts into batches. Based on the rollouts from each batch, we instruct $M_\text{skill}$ to \emph{refine} $s_\text{W}$ by adding \texttt{Principles} while preserving the \texttt{Workflow}. Since principles are list-like and merge-friendly, we hierarchically merge the locally refined skills of each branch into a global skill. Because the merging process introduces significant variation in global-skill performance, we additionally resample by shuffling and merging the refined local skills from the branch whose global skill performs best on $\mathcal{S}_\text{evo}$, thus obtaining $k_\text{smp}$ full skills $S_\text{Full}$. A concise workflow-only skill can sometimes outperform full skills, so we select the best-performing skill from the union of $S_\text{W}$ and $S_\text{Full}$ as our final skill.

\begin{table*}[!t]
\setlength{\belowcaptionskip}{-12pt}
\centering
\footnotesize
\begin{tblr}{
width=\linewidth,
colspec={X[2.1,l] X[1.9,l] *{6}{X[0.7,c]} *{5}{X[0.7,c]}},
colsep=1pt,
rowsep=1pt,
rows={valign=m},
row{1,2}={font=\bfseries},
column{9}={leftsep=5pt},
cell{1}{1}={r=2}{c,m},
cell{1}{2}={r=2}{c,m},
cell{1}{3}={c=6}{c},
cell{1}{9}={c=5}{c},
hline{1,Z}={1pt,solid},
hline{2}={3-8,9-13}{0.5pt,solid,endpos,lr},
hline{3,4,10,11,20,21,25,26}={0.6pt,solid},
}
{Judge~Model} & {Rubric~Model} & RewardBench~2 & & & & & & RewardBench & & & & \\
& & Fact. & Foc. & Math & PIF & Safe. & Avg. & Chat & Hard & Safe. & Rsn. & Avg. \\
\SetCell[c=13]{l} \textbf{Judge Backbone: Qwen3-4B and Variants} & & & & & & & & & & & & \\
Qwen3-4B & w/o & 48.80 & 74.77 & 57.43 & 26.67 & 71.33 & 55.80 & \uline{93.54} & 61.42 & 84.48 & 82.06 & 80.38 \\
Qwen3-4B & Rubric-RM-4B & 44.27 & 76.20 & 50.60 & 31.67 & 35.43 & 47.63 & 91.47 & 64.33 & 60.16 & 81.19 & 74.29 \\
Qwen3-4B & DSV4 & 45.78 & 76.79 & \textbf{63.86} & 34.44 & 51.33 & 54.44 & 91.99 & \textbf{70.88} & 82.55 & 83.56 & \uline{82.25} \\
Rubric-RM-4B & Rubric-RM-4B & 45.33 & \uline{78.73} & 56.63 & \textbf{41.67} & 34.29 & 51.33 & 91.09 & 67.98 & 46.72 & 81.05 & 71.71 \\
\SetRow{bg=gray!15} Q3-4B + Naive~Skill & w/o & \uline{57.96} & 72.41 & 59.44 & 38.33 & \uline{83.43} & \uline{62.31} & 92.12 & 52.06 & \uline{87.55} & \uline{83.86} & 78.90 \\
\SetRow{bg=gray!15} Q3-4B + Eval-Skill & w/o & \textbf{59.02} & \textbf{85.06} & \uline{62.25} & \uline{40.56} & \textbf{90.86} & \textbf{67.55} & \textbf{93.67} & \uline{70.79} & \textbf{88.23} & \textbf{89.40} & \textbf{85.52} \\
\SetCell[c=13]{l} \textbf{Judge Backbone: Qwen3-8B and Variants} & & & & & & & & & & & & \\
Qwen3-8B & w/o & 52.18 & 74.35 & 59.04 & 30.00 & 69.62 & 57.04 & \uline{94.44} & 64.61 & 86.41 & 84.09 & 82.39 \\
Qwen3-8B & Rubric-RM-8B & 43.12 & 76.71 & 56.63 & 37.78 & 42.38 & 51.32 & 88.76 & 68.82 & 65.62 & 77.14 & 75.09 \\
Qwen3-8B & Rubric-ARM-8B & 47.73 & 77.72 & 51.81 & 39.44 & 41.14 & 51.57 & 91.73 & 72.94 & 68.12 & 81.61 & 78.60 \\
Qwen3-8B & DSV4 & 50.49 & 77.38 & 62.65 & 41.11 & 53.43 & 57.01 & 91.73 & 71.35 & 83.49 & 84.31 & 82.72 \\
Rubric-RM-8B & Rubric-RM-8B & 43.90 & 84.98 & 61.85 & 43.89 & 37.24 & 54.37 & 92.25 & 74.16 & 57.66 & 83.61 & 76.92 \\
Rubric-ARM-8B & Rubric-ARM-8B & 51.79 & \uline{89.01} & 63.13 & \uline{57.33} & 42.69 & 60.79 & 91.32 & \uline{77.98} & 62.84 & 85.37 & 79.38 \\
\SetRow{bg=gray!15} Q3-8B + Naive~Skill & w/o & 54.22 & 73.33 & 63.05 & 33.33 & 90.67 & 62.92 & 93.02 & 67.98 & 88.02 & 85.01 & 83.51 \\
\SetRow{bg=gray!15} Q3-8B + Eval-Skill & w/o & \uline{64.27} & 80.51 & \textbf{71.49} & 42.78 & \uline{93.33} & \uline{70.48} & 93.67 & 72.66 & \textbf{89.11} & \textbf{92.43} & \uline{86.97} \\
\SetRow{bg=gray!15} Rubric-ARM-8B + Eval-Skill & w/o & \textbf{65.69} & \textbf{90.04} & \uline{67.47} & \textbf{89.89} & \textbf{94.67} & \textbf{75.35} & \textbf{95.09} & \textbf{80.24} & \uline{88.59} & \uline{89.80} & \textbf{88.43} \\

\SetCell[c=13]{l} \textbf{Judge Backbone: DeepSeek-V4-Flash} & & & & & & & & & & & & \\
DSV4 & w/o & 69.07 & 88.61 & 70.68 & 45.00 & 68.00 & 68.27 & \uline{94.96} & 80.06 & 88.12 & 95.49 & 89.66 \\
DSV4 & DSV4 & 64.98 & 87.43 & 73.49 & 56.67 & 60.86 & 68.69 & 91.34 & 82.02 & 83.28 & 95.36 & 88.00 \\
\SetRow{bg=gray!15} DSV4 + Naive Skill & w/o & \uline{80.71} & \uline{90.97} & \uline{83.53} & \uline{73.89} & \uline{95.43} & \uline{84.91} & 94.19 & \uline{84.55} & \textbf{93.02} & \uline{98.02} & \uline{92.45} \\
\SetRow{bg=gray!15} DSV4 + Eval-Skill & w/o & \textbf{84.36} & \textbf{92.32} & \textbf{84.74} & \textbf{76.11} & \textbf{96.38} & \textbf{86.78} & \textbf{95.87} & \textbf{87.55} & \uline{92.55} & \textbf{99.02} & \textbf{93.75} \\
\SetCell[c=13]{l} \textbf{Judge Backbone: Other Post-Trained RMs} & & & & & & & & & & & & \\
RM-R1-Qwen2.5-7B & w/o & 39.82 & 70.80 & 44.58 & 36.67 & 44.38 & 47.25 & 82.95 & 65.17 & 76.35 & 82.86 & 76.83 \\
RM-R1-DS-Qwen-7B & w/o & 36.36 & 64.39 & 61.45 & 21.67 & 31.52 & 43.08 & 79.33 & 66.67 & 74.06 & 90.10 & 77.54 \\
RRM-7B & w/o & 36.00 & 64.64 & 56.63 & 20.56 & 43.43 & 44.25 & 84.63 & 67.60 & 79.38 & 88.17 & 79.95 \\
RRM-32B & w/o & 55.38 & 78.90 & 76.71 & 35.00 & 55.14 & 60.23 & 94.70 & 73.31 & 85.00 & 97.79 & 87.70 \\
\end{tblr}

\caption{Main experiment results on RewardBench~2 and RewardBench. Within each judge group, the highest score is bolded and the second-highest score is underlined.}
\label{tab:main_results}
\end{table*}

\section{Experiments}
\label{sec:Experiment}

\subsection{Datasets and Experiment Settings}
\label{ssec:experiment_setting}
\paragraph{Datasets}
We conduct our main experiments on three multi-domain RM benchmarks: RewardBench~2, RewardBench, and RM-Bench. For each domain, we randomly sample 100 cases\footnote{The chat domain from RM-Bench has only 129 unique cases, so we use 50 cases for skill evolution.} as the evolving set for skill generation and reserve the remaining cases for testing. For RewardBench~2 and RewardBench, we report average@3, and for RM-Bench, we report the average over the $3 \times 3$ chosen--rejected grids. We further analyze \texttt{Eval-Skill}'s application to JudgeBench (\S~\ref{ssec:mixed}) and HealthBench (\S~\ref{ssec:hb}).

\paragraph{Backbones}
In the main experiments, we use Qwen3-4B, Qwen3-8B \citep{yang2025qwen}, DeepSeek-V4-Flash \citep{deepseek-ai2026deepseekv}, and the post-trained reward model Rubric-ARM-8B-Judge as $M_\text{judge}$ backbones. For each setting, we use the same backbone for both evolution rollout generation and judgment. We use DeepSeek-V4-Flash (with thinking enabled) as $M_\text{skill}$ for skill management, including generation, refinement, and merging. For an analysis of \texttt{Eval-Skill}'s applicability to other backbones, see \S~\ref{ssec:other_backbones}. We set $k_\text{gen}$ to 5, $k_\text{sel}$ to 3, and $k_\text{smp}$ to 5 for the main experiments, and we explore scaling these hyperparameters in \S~\ref{ssec:sets}.

\paragraph{Baselines}
For reward modeling, we compare against four types of baselines:

\begin{itemize}
    \item Vanilla one-step method: the judge model evaluates without any rubric or skill.
    \item Rubric-based methods: naive self-generation and post-trained methods, including Rubric-RM \citep{liu2025openrubrics} and Rubric-ARM \citep{xu2026alternating}, both of which include a post-trained rubric model and a post-trained RM.
    \item Post-trained RMs: RRM \citep{guo2025reward} and RM-R1 \citep{chen2025rmr}.
    \item Naive skill method: as introduced in \S~\ref{ssec:naive_skill_method}.
\end{itemize}

Additional implementation details are provided in Appendix~\ref{apx:implementation_details}. We further compare Eval-Skill with Skill-RM, a concurrent skill-based method, in Appendix~\ref{aapx:comparison_with_skill_rm}.

\subsection{Main Experiment Results}
\label{ssec:main_experiment_results}
We report the results on RewardBench~2 and RewardBench in Table~\ref{tab:main_results}, and the results on RM-Bench in Table~\ref{tab:main_rmbench} in Appendix~\ref{aapx:main_experiment}.

\paragraph{Strong Performance across Models}
As shown in Table~\ref{tab:main_results}, \texttt{Eval-Skill} outperforms all baseline types within each judge group, surpassing the vanilla one-step method by more than 10\% on RewardBench~2 and \~5\% on RewardBench. Although DeepSeek-V4-Flash itself serves as the skill manager, it is still notably improved by \texttt{Eval-Skill}, suggesting that \texttt{Eval-Skill}'s gains are not bounded by the skill manager's standalone ability. These results demonstrate that \texttt{Eval-Skill} can substantially improve RM performance across models with diverse capabilities and across diverse domains.

\paragraph{Further Improvement on Post-trained RMs}
\texttt{Eval-Skill} is plug-and-play and can be applied directly to post-trained RMs. When applied to Rubric-ARM-8B-Judge, it improves the RewardBench~2 average by 13.22 percentage points, outperforming both the direct Rubric-ARM-8B-Judge and Qwen3-8B+\texttt{Eval-Skill} settings. This result shows that \texttt{Eval-Skill} can further boost post-trained models.

\paragraph{Superiority over the Naive Skill Method}
Although the naive skill method already demonstrates strong performance, \texttt{Eval-Skill} further improves it, especially on weaker models such as Qwen3-4B and Qwen3-8B. This highlights the importance of skill quality and shows that \texttt{Eval-Skill} can substantially improve the robustness of skill generation.

\subsection{Skill Evolution-Time Scaling}
\label{ssec:sets}
We define \textbf{Skill Evolution-Time Scaling} (SETS) as a paradigm for improving a skill-based method's performance by allocating additional compute during the skill evolution phase. Because skill quality is critical and compute spent during skill evolution typically does not add inference-time overhead, SETS can be an efficient way to improve performance.

We conduct both sequential and parallel SETS experiments. In sequential scaling, the skill is iteratively refined based on rollouts generated using the previous-round skill. In parallel scaling, we control the sample size used to generate multiple skills in \texttt{Eval-Skill} before choosing the best one according to performance on the evolving set.

\subsubsection{Sequential Scaling}
\label{sssec:sequential_scaling}
\begin{figure*}[!t]
\setlength{\abovecaptionskip}{5pt}
\setlength{\belowcaptionskip}{-12pt}
    \centering
    \includegraphics[width=\linewidth]{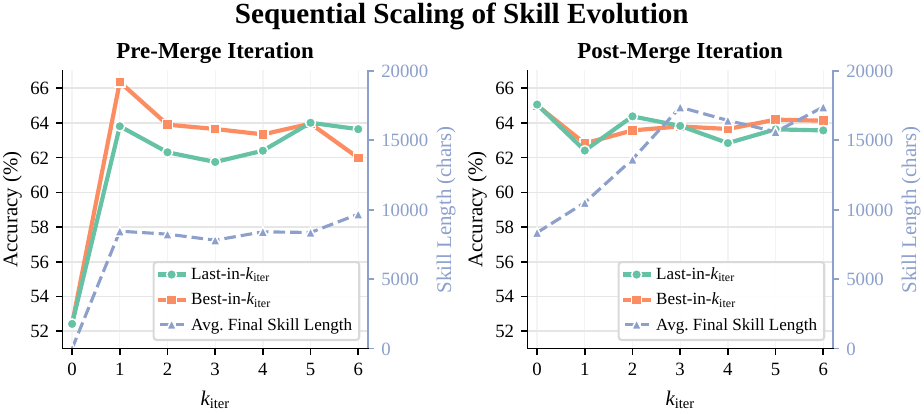}
    \caption{Sequential scaling of skill evolution on RewardBench~2 under pre-merge and post-merge iteration. For post-merge iteration, we set the pre-merge $k_\text{iter}$ is set to 1 (best-in-1). Accuracy is measured on the test set using the merged skill.}
    \label{fig:sequential_scaling}
\end{figure*}

For sequential scaling, we iteratively refine the skill using rollouts guided by the skill from the previous round, starting from no skill (for pre-merge $k_\text{iter}$=0). we use the one-stage setting and remove first-round supervision (as in \S~\ref{ssec:ablation}) to isolate the effect of iteration. We vary the number of refinement rounds $k_\text{iter}$ from 0 to 6 for both pre-merge and post-merge iteration. In contrast to observations in other domains, \textbf{we find that, for reward modeling, iterative skill refinement does not produce consistent improvement}. As shown in Figure~\ref{fig:sequential_scaling}, accuracy quickly reaches a similar range and then fluctuates rather than increasing monotonically.

By analyzing generated skills across iterations, we attribute this behavior to two factors: (1) for reward modeling tasks, preference alignment typically matters more than accumulating additional task knowledge, so creating a workable workflow is not difficult, but finding a better-aligned one is; and (2) selecting iterative refinements on a small evolving set can overfit to that set. More details on the experimental setting and discussion are provided in Appendix~\ref{aapx:sequential_scaling}.

\subsubsection{Parallel Scaling}
\label{sssec:parallel_scaling}
\begin{figure*}[!t]
\setlength{\abovecaptionskip}{5pt}
\setlength{\belowcaptionskip}{-12pt}
    \centering
    \includegraphics[width=\linewidth]{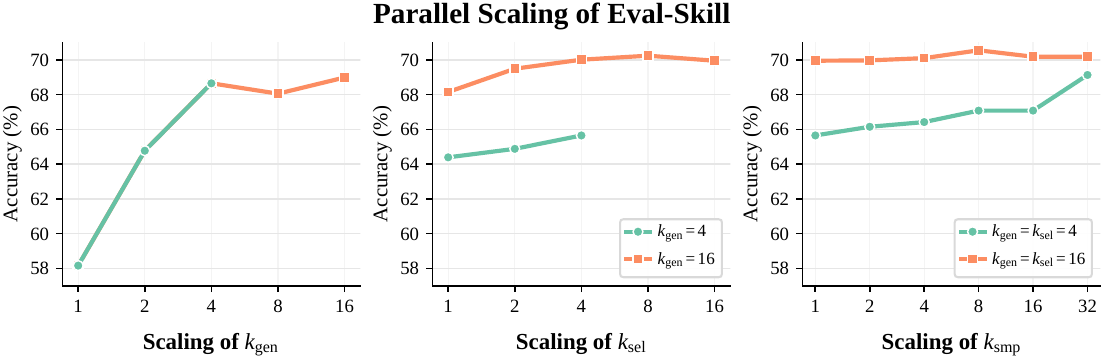}
    \caption{Scaling the hyperparameters $k_\text{gen}$, $k_\text{sel}$, and $k_\text{smp}$ in \texttt{Eval-Skill} shows that more samples lead to better performance, although performance eventually plateaus. Accuracy is measured on the test set using the chosen skill after each sampling round.}
    \label{fig:parallel_scaling}
\end{figure*}

For parallel scaling, we use the same setting as \texttt{Eval-Skill} (with the crossover replaced with an additionaly selection) and scale the three hyperparameters $k_\text{gen}$, $k_\text{sel}$, and $k_\text{smp}$. The results in Figure~\ref{fig:parallel_scaling} show two trends: (1) scaling all three hyperparameters can improve performance, with $k_\text{gen}$ having the largest effect; and (2) when $k_\text{gen}$ is already high, the gains from scaling $k_\text{sel}$ and $k_\text{smp}$ are small (from 69.95\% for $k_\text{smp}=1$ to 70.18\% for $k_\text{smp}=32$), because performance quickly plateaus, indicating a soft upper bound. More details on the experimental setting are provided in Appendix~\ref{aapx:parallel_scaling}.

\subsection{Reward-Guided Best-of-N Performance}
\label{ssec:bon}
\begin{figure*}[!t]
\setlength{\abovecaptionskip}{5pt}
\setlength{\belowcaptionskip}{-12pt}
    \centering
    \includegraphics[width=\linewidth]{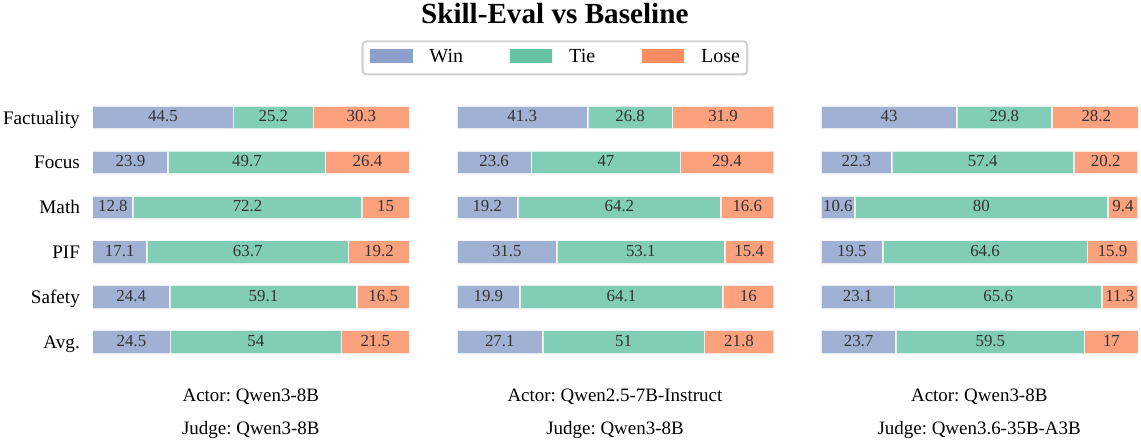}
    \caption{Reward-guided best-of-$N$ inference. Win, tie, and lose indicate the comparison between the \texttt{Eval-Skill}-guided verifier and the baseline verifier, as judged by DeepSeek-V4-Pro (with thinking enabled).}
    \label{fig:bon}
\end{figure*}

The preceding experiments use responses provided by the benchmarks. Here, following RRM \citep{guo2025reward}, we use \texttt{Eval-Skill} as a verifier to guide best-of-$N$ inference for an actor model. This setting serves as a proxy task for evaluating \texttt{Eval-Skill}'s ability to serve as an RM for reinforcement learning. For each case, we instruct an actor model to generate $N=8$ candidate responses. We then ask judges, with and without \texttt{Eval-Skill}, to select the best response through hierarchical pairwise selection. When the two settings choose different responses, we ask a final judge, DeepSeek-V4-Pro, to select a winner from their chosen responses or mark a tie when neither is clearly better. Figure~\ref{fig:bon} shows that \texttt{Eval-Skill} generally achieves a higher win rate than the baseline, although some settings and domains are more challenging. We attribute these weaker cases to the relative scarcity of substantive distinctions between candidate responses and to the mismatch between the task preference encoded in the skill and the preference of the final judge, which is guided only by a one-sentence domain description to avoid bias.

\section{Analyses and Discussion}
\label{sec:Analyses}

\subsection{How does \texttt{Eval-Skill} work with other backbones and across backbones?}
\label{ssec:other_backbones}
\paragraph{Performance on Other Backbones}
To further test \texttt{Eval-Skill}'s generalizability, we evaluate it on three additional types of backbone models: (1) models from other families, including Llama-3.1-8B-Instruct, Qwen3.6-35B-A3B, and Gemma-4-E4B-IT; (2) Qwen3-8B with thinking mode enabled; and (3) another post-trained generative RM, RRM-7B. Results in Tables~\ref{tab:other_backbones1} and \ref{tab:other_backbones2} show that \texttt{Eval-Skill} generalizes well across backbones.

As discussed in Appendix~\ref{aapx:performance_on_other_backbones}, we also observe that models with weaker instruction-following ability, including Llama-3.1-8B-Instruct and RRM-7B, may not follow skills reliably. For these models, a shorter, workflow-only skill can perform better than a full skill.

\paragraph{Transferability Across Backbones}
We test whether \texttt{Eval-Skill} skills transfer across backbone models. As shown in Table~\ref{tab:trans_model_all}, skills generated from rollouts of one model can bring similar gains to models with different capabilities, and they significantly outperform the no-skill baseline. This highlights the efficiency of \texttt{Eval-Skill}: generated skills are often model-transferable, so a skill evolved from one model's rollouts can be applied to multiple models.

\subsection{How do skills work across domains and in mixed-domain settings?}
\label{ssec:mixed}

\paragraph{Transferability Across Domains}
We heuristically construct two domain groups based on potential domain similarity: the Chat Group and the Reasoning Group. We use these groups to investigate skill transfer across domains. Table~\ref{tab:trans_domain} reports OOD performance relative to the no-skill baselines. Performance is strongly influenced by the match between the source and target domains. For example, we observe noticeable transferability among the Chat domain from RewardBench, the Chat domain from RM-Bench, and the Factuality domain from RewardBench~2, whereas mismatched skills can significantly degrade performance in other cases. This suggests that skills capture domain-specific evaluation priorities rather than acting as universally beneficial prompts.

\begin{table*}[!t]
\centering
\footnotesize
\begin{minipage}[t]{0.57\linewidth}
\centering
\setlength{\tabcolsep}{2pt}
\renewcommand{\arraystretch}{1.08}
\begin{tabular}{@{}>{\bfseries}l *{5}{c}@{}}
\Xhline{1pt}
\makecell[l]{Target\textbackslash{}Source} & \bfseries RB.Chat & \bfseries RB.CH & \bfseries RMB.Chat & \bfseries RB2.Fact & \bfseries RB2.Focus \\
\Xhline{0.6pt}
RB.Chat & \cellcolor[HTML]{FEEA83}-0.82 & \cellcolor[HTML]{FA9B74}-13.27 & \cellcolor[HTML]{FDD27F}-4.51 & \cellcolor[HTML]{FCB679}-9.02 & \cellcolor[HTML]{FA9573}-14.22 \\
RB.CH & \cellcolor[HTML]{FEEB84}-0.43 & \cellcolor[HTML]{AAD380}{\uline{12.46}} & \cellcolor[HTML]{E0E383}{\uline{4.06}} & \cellcolor[HTML]{D4DF82}{\uline{5.94}} & \cellcolor[HTML]{F0E784}{\uline{1.73}} \\
RMB.Chat & \cellcolor[HTML]{FCBD7B}-7.84 & \cellcolor[HTML]{ABD380}{\uline{12.25}} & \cellcolor[HTML]{B2D580}{\uline{11.13}} & \cellcolor[HTML]{B2D580}{\uline{11.13}} & \cellcolor[HTML]{FDCD7E}-5.39 \\
RB2.Fact & \cellcolor[HTML]{FCC57C}-6.65 & \cellcolor[HTML]{FFEB84}-0.69 & \cellcolor[HTML]{A1D07F}{\uline{13.80}} & \cellcolor[HTML]{63BE7B}23.17 & \cellcolor[HTML]{F8696B}-21.29 \\
RB2.Focus & \cellcolor[HTML]{FEE182}-2.17 & \cellcolor[HTML]{E7E483}{\uline{3.05}} & \cellcolor[HTML]{FBB279}-9.66 & \cellcolor[HTML]{FCBE7B}-7.72 & \cellcolor[HTML]{C5DB81}8.29 \\
\Xhline{0.6pt}
\end{tabular}
\end{minipage}
\hfill
\begin{minipage}[t]{0.42\linewidth}
\centering
\setlength{\tabcolsep}{0.8pt}
\renewcommand{\arraystretch}{1.15}
\begin{tabular}{@{}>{\bfseries}l *{3}{c}@{}}
\Xhline{1pt}
\makecell[l]{Target\textbackslash{}Source} & \makecell{\bfseries RB.\\\bfseries Reason.} & \makecell{\bfseries RB2.\\\bfseries Math} & \makecell{\bfseries RMB.\\\bfseries Math} \\
\Xhline{0.6pt}
RB.Reason. & \cellcolor[HTML]{FEEF9C}9.92 & \cellcolor[HTML]{FFEF9C}9.74 & \cellcolor[HTML]{FFEF9C}9.80 \\
\hdashline
RB2.Math & \cellcolor[HTML]{B1D78C}21.09 & \cellcolor[HTML]{B1D78C}21.09 & \cellcolor[HTML]{F8ED9B}10.87 \\
\hdashline
RMB.Math & \cellcolor[HTML]{63BE7B}32.26 & \cellcolor[HTML]{D3E293}16.15 & \cellcolor[HTML]{CDE092}16.99 \\
\Xhline{0.6pt}
\end{tabular}
\end{minipage}

\caption{Cross-domain transfer performance relative to the no-skill baselines. Left: the Chat Group. Right: the Reasoning Group.}
\label{tab:trans_domain}
\end{table*}

\begin{figure}[!t]
\setlength{\abovecaptionskip}{5pt}
\setlength{\belowcaptionskip}{-12pt}
    \centering
    \includegraphics[width=\linewidth]{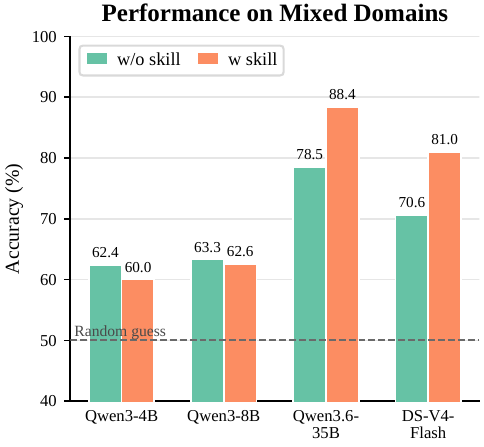}
    \caption{Performance on the mixed-domain JudgeBench setting.}
    \label{fig:mixed_domain}
\end{figure}

\paragraph{Performance on Mixed Domains}
We test \texttt{Eval-Skill} on JudgeBench, an RM benchmark with four domains: general knowledge, reasoning, mathematics, and coding. Instead of creating one skill per domain as in the main experiments (see \S~\ref{ssec:experiment_setting}), we mix cases from different domains and generate a single mixed-domain skill. The results in Figure~\ref{fig:mixed_domain} show that while this skill can still benefit stronger models, it does not improve the performance of weaker models such as Qwen3-4B and Qwen3-8B, and can even slightly decrease it.

By inspecting the generated skills, we find that the workflow part typically contains branching structures tailored to different task domains (e.g., see Figure~\ref{fig:skill_q4_jb}), showing the adaptability of the skill generation process. We attribute the limited improvement on weaker models to their difficulty in following the complex branching and conditions specified in the skills.

\FloatBarrier

\subsection{A Stress Test on HealthBench}
\label{ssec:hb}
To test \texttt{Eval-Skill} on more challenging tasks and settings, we conduct experiments on HealthBench, a benchmark measuring the performance and safety of LLMs in healthcare. Because HealthBench is not an RM benchmark, we adapt it by iteratively generating responses with Qwen3-8B for each query, scoring them with the ground-truth rubrics, and retaining a pair once two responses for the same query differ by at least 0.1. We evaluate judge models by providing the prompt and two responses without the rubrics. The results in Table~\ref{tab:app_hb} indicate that, except for DeepSeek-V4-Flash, \texttt{Eval-Skill} does not yield significant gains. We attribute this to the knowledge-intensive nature of HealthBench and the intrinsic difficulty of discriminating between two Qwen3-8B responses.

\subsection{Ablation Experiments}
\label{ssec:ablation}

\paragraph{Alternative Single-Stage Recipes}
\texttt{Eval-Skill} is designed as a two-stage method: it first generates a workflow and then derives principles. We also test alternative one-stage settings: (1) \emph{workflow-only}, i.e., the stage-1-only version of \texttt{Eval-Skill}; (2) \emph{principle-only}, i.e., the stage-2-only version of \texttt{Eval-Skill}; and (3) \emph{full}, which directly generates a full skill in one stage. The results in Table~\ref{tab:ablation_stage} show that these variants typically achieve similar performance, only slightly lagging behind the full \texttt{Eval-Skill} method. This demonstrates that both the workflow component and the principle component are highly expressive for guiding judges, while sharing certain expressive capabilities, such as the ability to encode conditional logic through if-clauses.

\paragraph{Others}
We also verify the effects of (1) first-round supervision and (2) reference-guided generation in stage~1, as shown in Table~\ref{tab:ablation_stage}.

\begin{table}[!ht]
\centering
\footnotesize
\begin{tblr}{
width=\linewidth,
colspec={X[2.2,l] *{4}{X[0.85,c]}},
colsep=3pt,
rowsep=1pt,
rows={valign=m},
row{1,2}={font=\bfseries},
cell{1}{1}={r=2}{c,m},
cell{1}{2}={c=2}{c},
cell{1}{4}={c=2}{c},
cell{3,7}{1}={c=5}{l,font=\itshape},
hlines={0.25pt,dashed},
hline{1,Z}={1pt,solid},
hline{2}={2-3,4-5}{0.5pt,solid,endpos,lr},
hline{3,4,7,8}={0.3pt,solid},
}
Method & Qwen3-4B & & Qwen3-8B & \\
& RB-2 & RB & RB-2 & RB \\
Two-Stage Paradigm & & & & \\
Eval-Skill & \textbf{67.55} & \textbf{85.52} & \textbf{70.48} & 86.97 \\
\quad{}-- first-round supervision & 65.38 & 81.84 & 70.41 & 85.37 \\
\quad{}-- stage 1 exploration & 62.55 & 83.12 & 66.96 & 83.38 \\
Single-Stage Paradigm & & & & \\
workflow-only & 61.28 & 82.96 & 68.05 & 83.80 \\
principle-only & 66.59 & 84.25 & 69.84 & \textbf{87.21} \\
full & 63.97 & 80.89 & 65.55 & 85.19 \\
\end{tblr}

\caption{Ablation results for the two-stage design and alternative single-stage recipes on RewardBench~2 and RewardBench.}
\label{tab:ablation_stage}
\end{table}
\FloatBarrier

\section{Related Work}
\label{sec:RW}
\paragraph{Rubric-Based Reward Modeling}
Recent reward modeling has expanded from scalar reward prediction toward generative judging, especially for open-ended tasks where verifiable answers are unavailable. In this setting, rubrics and checklists provide explicit evaluation standards for LLM-as-a-judge (LAAJ) systems, much like human teachers grading exam answers against predefined criteria \citep{kim2024prometheus}. A rubric or checklist typically consists of multiple criteria and can be specified at the task level or case level. Several benchmarks provide human-crafted case-level rubrics to support LAAJ evaluation \citep{arora2025healthbench,sharma2025researchrubrics,starace2025paperbench}.

Online rubric generation has also been used as an intermediate step in reward modeling systems, where a rubric model first generates evaluation criteria and a downstream judge then applies them. When reference answers are available, they can be incorporated into rubric generation to improve reward modeling for reinforcement learning \citep{gunjal2025rubrics,huang2025reinforcement,liu2025openrubrics}. In many open-ended settings, however, reference answers are unavailable, so the rubric is generated from the user query alone \citep{xu2026alternating,he2025advancedif,dou2025baichuanm,dhole2026rubricrag}. RubricBench \citep{zhang2026rubricbench} further evaluates the ability of models to generate rubrics, highlighting rubric self-generation as a standalone capability. Although generated rubrics can in principle reduce surface-level biases and help judges attend to task-specific criteria, our results show that they can also hurt judge performance and even underperform vanilla baselines. RIFT \citep{qi2026rift} provides a taxonomy of rubric failure modes; building on this perspective, we empirically verify related failures on RM benchmarks and extend the representational space from rubrics to skills, which leads to stronger RM performance.

\paragraph{Skill-Based Methods}
Prior work has studied how experiences or insights extracted from rollouts on related cases can guide LLM behavior \citep{wang2023voyager,zhao2024expel,fang2025memp,ouyang2025reasoningbank,wu2025evolver}. Since Anthropic introduced Skills \citep{anthropic_skills}, skill-based methods have emerged across domains, including agents \citep{xia2026skillrl,jiang2026xskill,ouyang2026skillos}, mathematics \citep{li2026arise}, coding \citep{wang2026effiskill}, routing \citep{wang2026skillorchestra}, and LLM memory management \citep{zhang2026memskill}. Some methods use public or human-crafted skills \citep{zheng2026skillrouter,chen2026skillrm}, and SkillsBench \citep{li2026skillsbench} provides a benchmark for this paradigm. Other methods automatically extract domain-specific skills from rollouts on an evolving set \citep{xia2026skillrl,ni2026traceskill,jiang2026xskill,tu2026dynamic,wang2026skillx,zhang2026coevoskills}, often finding that automatically generated skills outperform human-curated ones \citep{ni2026traceskill,zhang2026coevoskills}.

Existing skill-based systems differ in how they store and deploy skills. Some maintain a skill library and retrieve relevant skills at inference time \citep{xia2026skillrl}, whereas others retain a single domain-level skill and inject it directly into the context of the executor model \citep{ni2026traceskill}. Because retrieval quality is a major bottleneck for skill-library methods, strongly affecting downstream performance \citep{zheng2026skillrouter}, and because additional retriever models introduce extra overhead and complexity, we adopt the \emph{one global skill} pattern in this work. We further emphasize the role of exploration and selection in RM skill generation: for preference-sensitive evaluation tasks, different workflows can encode different judgment priorities, so simply accumulating or iterating skills is not always sufficient. The most closely related works are two concurrent skill-based evaluation methods: RewardHarness \citep{zhang2026rewardharness} and Skill-RM \citep{chen2026skillrm}. RewardHarness focuses on image editing, whereas we study general RM scenarios across diverse domains. Skill-RM is closer in scope, but it relies on manually crafted RM skills and emphasizes agentic orchestration over a resource bank. In contrast, \texttt{Eval-Skill} emphasizes automatic skill synthesis and uses a relatively simple skill system, reducing overhead and placing lower demands on the backbone model. We also make a performance comparison with Skill-RM on Qwen3.5-27B, and find Eval-Skill generally outperforms Skill-RM (see Appendix~\ref{aapx:comparison_with_skill_rm}).

\section{Conclusion}
\label{sec:Conclusion}
We study the limitations of rubric-based reward modeling and show that online rubric generation can degrade reward model performance due to rigid criteria and misalignment with reward model capabilities. To address these issues, we explore an alternative paradigm that directly extracts reusable evaluation skills from model rollouts, without relying on online rubric generation. We further introduce \texttt{Eval-Skill}, which develops high-quality domain-level evaluation skills through two-stage progressive generation with exploration and selection interleaved across both workflow and principle construction. Experiments and analyses show that high-quality skills can consistently improve reward models and transfer across backbones and relevant domains, providing an efficient alternative to two-step rubric-based evaluation.


\section*{Limitations}
\label{sec:Limitations}
First, although we explored pairwise and listwise reward modeling, we did not include pointwise RM tasks into our scope. Second, while we validated \texttt{Eval-Skill} on reward modeling benchmarks and best-of-N inference, its effectiveness for downstream reinforcement learning remains to be verified. Third, as discussed in \S~\ref{ssec:mixed} and \S~\ref{ssec:hb}, mixed-domain settings and more challenging tasks can remain difficult, especially for smaller backbones.

\section*{Ethical Considerations}
\label{sec:ethics_statement}
\texttt{Eval-Skill} is designed to improve the reliability of reward-model-based evaluation. However, stronger evaluators may also reinforce objectives, biases, or omissions present in their training data, evolution sets, and skill prompts. When applying \texttt{Eval-Skill} to safety-critical or socially sensitive domains, practitioners should audit both the selected skills and the resulting reward model outputs, and should not treat skill-guided judgments as substitutes for expert human review.

\paragraph{Artifacts and Licenses}
We use publicly available research artifacts, including benchmark datasets, baseline methods, and open-source models, in accordance with their original licenses and terms of use. All artifacts are used solely for research and evaluation purposes. We will release our code, prompts, and generated skills under the Apache-2.0 license, and provide references to the original sources and licenses of all external artifacts used in our experiments.

\paragraph{Use of AI Assistants}
We used AI assistants, including ChatGPT, to support writing polishing, code and debugging assistance, and illustration generation. The authors reviewed, edited, and verified all AI-assisted outputs, and take full responsibility for the final content, experiments, analysis, and claims in this paper.

\bibliography{custom, custom_add}

\appendix
\label{appendix}
\section{Limitation of Rubric-Based Methods}
\label{apx:weakness_rubric}

\subsection{Detailed Comparison between Rubric-Based Methods and Baselines}
\label{aapx:detailed_comparison_rubric}
The detailed comparison between rubric-based methods and baselines on RewardBench~2 and RewardBench are shown in Table~\ref{tab:app_rubric_detail}.

\begin{table*}[!t]
\centering
\scriptsize
\begin{tblr}{
width=\linewidth,
colspec={X[1.35,l] X[1.35,l] *{6}{X[0.62,c]} *{5}{X[0.62,c]}},
colsep=1.4pt,
rowsep=1pt,
hline{1,Z}={1pt,solid},
hline{2}={3-8,9-13}{0.5pt,solid,endpos,lr},
hline{3,7,9,14,16,18,21,24}={0.5pt,solid},
row{1}={font=\bfseries},
row{2}={font=\bfseries,belowsep=1pt,valign=m},
column{9}={leftsep=5pt},
cell{1}{1}={r=2}{c,m},
cell{1}{2}={r=2}{c,m},
cell{1}{3}={c=6}{c},
cell{1}{9}={c=5}{c},
}
Judge Model & Rubric Model & RewardBench 2 & & & & & & RewardBench & & & & \\
& & Fact. & Focus & Math & PIF & Safety & Avg. & Chat & Chat-Hard & Safety & Rsn. & Avg. \\
Qwen3-4B & w/o & \textbf{48.80} & 74.77 & 57.43 & 26.67 & \textbf{71.33} & \textbf{55.80} & \textbf{93.54} & 61.42 & \textbf{84.48} & 82.06 & 80.38 \\
Qwen3-4B & Qwen3-4B & 42.22 & 73.84 & 48.59 & 32.22 & 33.62 & 46.10 & 92.76 & 65.54 & 64.58 & 82.31 & 76.30 \\
Qwen3-4B & Rubric-RM-4B & 44.27 & 76.20 & 50.60 & 31.67 & 35.43 & 47.63 & 91.47 & 64.33 & 60.16 & 81.19 & 74.29 \\
Qwen3-4B & DS-V4-Flash & 45.78 & \textbf{76.79} & \textbf{63.86} & \textbf{34.44} & 51.33 & 54.44 & 91.99 & \textbf{70.88} & 82.55 & \textbf{83.56} & \textbf{82.25} \\
Rubric-RM-4B & w/o & \textbf{50.93} & \textbf{79.07} & \textbf{56.63} & \textbf{43.33} & \textbf{52.48} & \textbf{56.49} & \textbf{94.57} & 66.76 & \textbf{73.12} & 80.45 & \textbf{78.73} \\
Rubric-RM-4B & Rubric-RM-4B & 45.33 & 78.73 & \textbf{56.63} & 41.67 & 34.29 & 51.33 & 91.09 & \textbf{67.98} & 46.72 & \textbf{81.05} & 71.71 \\
Qwen3-8B & w/o & \textbf{52.18} & 74.35 & 59.04 & 30.00 & \textbf{69.62} & \textbf{57.04} & \textbf{94.44} & 64.61 & \textbf{86.41} & 84.09 & 82.39 \\
Qwen3-8B & Qwen3-8B & 45.28 & 73.97 & 58.07 & 34.67 & 41.14 & 50.63 & 90.70 & 69.01 & 70.89 & 81.43 & 78.01 \\
Qwen3-8B & Rubric-RM-8B & 43.12 & 76.71 & 56.63 & 37.78 & 42.38 & 51.32 & 88.76 & 68.82 & 65.62 & 77.14 & 75.09 \\
Qwen3-8B & Rubric-ARM-8B & 47.73 & \textbf{77.72} & 51.81 & 39.44 & 41.14 & 51.57 & 91.73 & \textbf{72.94} & 68.12 & 81.61 & 78.60 \\
Qwen3-8B & DS-V4-Flash & 50.49 & 77.38 & \textbf{62.65} & \textbf{41.11} & 53.43 & 57.01 & 91.73 & 71.35 & 83.49 & \textbf{84.31} & \textbf{82.72} \\
Rubric-RM-8B & w/o & \textbf{50.67} & 82.95 & 57.83 & 42.22 & \textbf{45.62} & \textbf{55.86} & \textbf{92.89} & 71.07 & \textbf{76.61} & \textbf{85.09} & \textbf{81.42} \\
Rubric-RM-8B & Rubric-RM-8B & 43.90 & \textbf{84.98} & \textbf{61.85} & \textbf{43.89} & 37.24 & 54.37 & 92.25 & \textbf{74.16} & 57.66 & 83.61 & 76.92 \\
Rubric-ARM-8B & w/o & \textbf{56.53} & 88.02 & 58.23 & 45.00 & \textbf{62.29} & \textbf{62.01} & \textbf{93.67} & 77.25 & \textbf{84.53} & \textbf{86.37} & \textbf{85.46} \\
Rubric-ARM-8B & Rubric-ARM-8B & 51.79 & \textbf{89.01} & \textbf{63.13} & \textbf{57.33} & 42.69 & 60.79 & 91.32 & \textbf{77.98} & 62.84 & 85.37 & 79.38 \\
Qwen3-8B-think & w/o & \textbf{57.60} & \textbf{82.87} & \textbf{85.14} & \textbf{46.67} & \textbf{65.62} & \textbf{67.58} & \textbf{93.80} & 81.18 & \textbf{87.55} & \textbf{97.94} & \textbf{90.12} \\
Qwen3-8B-think & Qwen3-8B-think & 52.80 & 80.42 & 79.12 & \textbf{46.67} & 43.90 & 60.58 & 91.60 & 79.12 & 68.07 & 97.09 & 83.97 \\
Qwen3-8B-think & DS-V4-Flash & 55.11 & 80.51 & 79.12 & 41.67 & 56.29 & 62.54 & 91.34 & \textbf{81.93} & 85.05 & 97.39 & 88.93 \\
Qwen3.6-35B-A3B & w/o & \textbf{75.29} & \textbf{85.32} & \textbf{85.14} & 63.33 & \textbf{77.14} & \textbf{77.24} & \textbf{93.02} & \textbf{83.15} & \textbf{91.98} & 98.67 & \textbf{91.71} \\
Qwen3.6-35B-A3B & Qwen3.6-35B-A3B & 71.20 & 82.03 & 84.74 & 67.78 & 73.24 & 75.80 & 91.09 & 82.49 & 88.75 & \textbf{98.77} & 90.28 \\
Qwen3.6-35B-A3B & DS-V4-Flash & 68.89 & 81.69 & 84.74 & \textbf{73.89} & 64.95 & 74.83 & 89.53 & 82.58 & 87.03 & 98.62 & 89.44 \\
DS-V4-Flash & w/o & \textbf{69.07} & \textbf{88.61} & 70.68 & 45.00 & \textbf{68.00} & 68.27 & \textbf{94.96} & 80.06 & \textbf{88.12} & \textbf{95.49} & \textbf{89.66} \\
DS-V4-Flash & DS-V4-Flash & 64.98 & 87.43 & \textbf{73.49} & \textbf{56.67} & 60.86 & \textbf{68.69} & 91.34 & \textbf{82.02} & 83.28 & 95.36 & 88.00 \\
\end{tblr}

\caption{Detailed comparison between rubric-based methods and baselines on RewardBench~2 and RewardBench. The highest score within each judge group is bolded for each column.}
\label{tab:app_rubric_detail}
\end{table*}

\subsection{Rubric Failure Mode Analysis}
\label{aapx:rubric_failure_mode_analysis}
\begin{figure*}[!t]
\setlength{\abovecaptionskip}{5pt}  
\setlength{\belowcaptionskip}{-12pt}
    \centering
    \includegraphics[width=\linewidth]{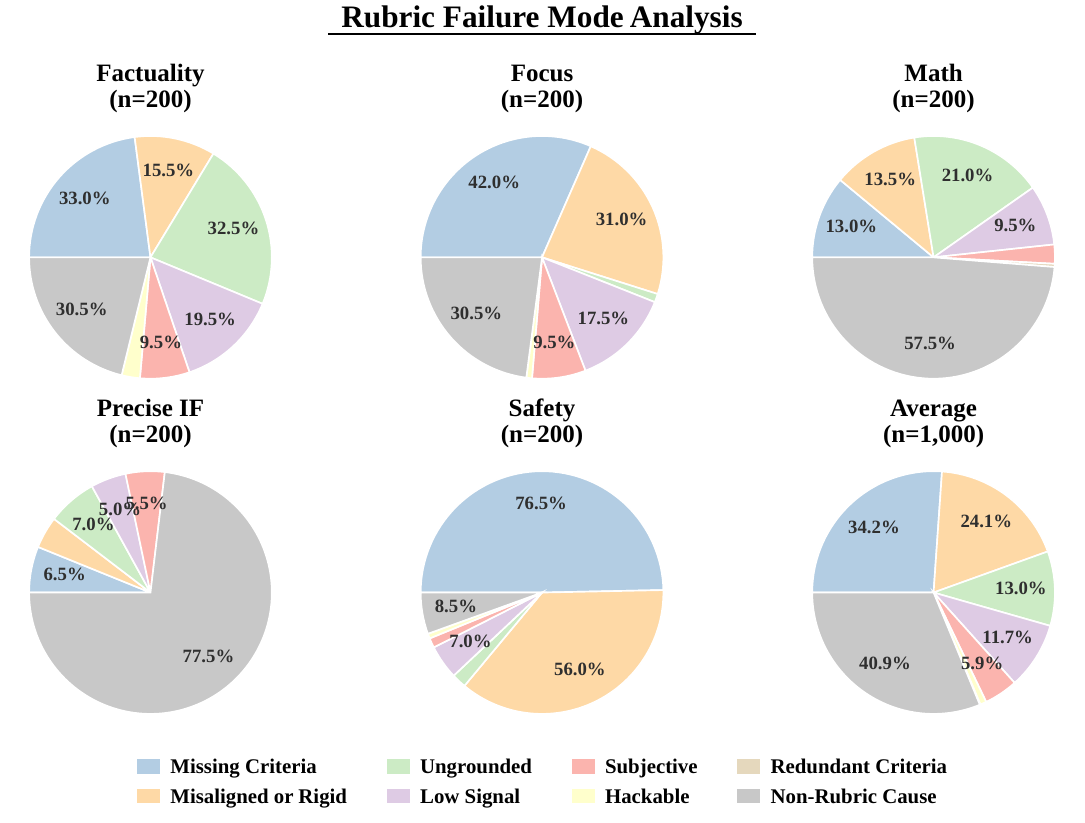}
    \caption{Rubric failure mode distribution on RewardBench~2, following the taxonomy of RIFT, with an additional non-rubric cause category.}
    \label{fig:rubric_failure_modes}
\end{figure*}
The results in Figure~\ref{fig:rubric_failure_modes} show that ``Missing Criteria'' (34.2\%) and ``Misaligned or Rigid'' (24.1\%) are frequent rubric-related failure modes. These errors are especially prominent in Factuality, Focus, and Safety, where response-agnostic rubrics often miss the decisive response-level distinction or impose criteria that do not match the candidates. For Math and Precise IF, many failures fall into the non-rubric category: even when the rubric asks for verification, the judge can still fail to verify the answer correctly.


\section{Eval-Skill Details}
\label{apx:eval_skill_details}

For local workflow-only skill synthesis, we generate rollouts with 5 trials per case and divide the cases into batches of size 4. For each batch, we independently synthesize local workflow-only skills in two rounds. In the first round, we use a strong model, $M_\text{supervision}$, instantiated as DeepSeek-V4-Flash with thinking enabled, to generate rollouts on the evolving set $\mathcal{S}*\text{evo}$ without any skill guidance. $M*\text{skill}$ then synthesizes local skills from these rollouts. In the second round, $M_\text{judge}$ generates rollouts on $\mathcal{S}*\text{evo}$ guided by the skill generated in the previous round, and $M*\text{skill}$ further refines the skill based on these rollouts. For each batch, we select the best-performing skill, or no skill if it performs better. We then provide these local skills to $M_\text{skill}$ to generate diverse new global skills, which are subsequently passed through selection.

For principle augmentation, we follow a similar procedure. For each branch, we generate rollouts with 5 trials per case and divide the cases into batches of size 4, using the workflow-only skill of that branch. For each batch, we independently synthesize full skills in two rounds. In each round, we refine the skill from the previous round and finally select the best-performing skill. Since these refinements are based on local rollouts, the resulting skills become local again. We therefore apply hierarchical merging, also with a batch size of 4, to obtain a global skill for each branch.

For additional skill samples, we reuse the local full skills from the best-performing branch, reshuffle them, and merge them to obtain multiple global skills. This incurs lower overhead than generating rollouts and iteratively refining skills, but also results in lower variance.

\section{Implementation Details}
\label{apx:implementation_details}

\subsection{Benchmark Details}
\label{aapx:benchmarks}
\paragraph{RewardBench~2} RewardBench~2 \citep{malik2025rewardbench} is a benchmark designed to provide new and challenging data for accuracy-based reward model evaluation. It contains 1,865 cases, divided into six subsets: Factuality, Focus, Math, Precise Instruction Following, Safety, and Tie. Except for the Tie subset, each case contains \textbf{a prompt, one chosen response, and three rejected responses}, generated by multiple LLMs.

Because the Tie subset does not fit our evaluation setting, we exclude it, leaving 1,763 cases. We split 100 cases from each of the five remaining subsets as the evolving set and use the rest as the test set. We provide judges with the prompt and the shuffled four responses, optionally accompanied by a skill, and ask them to choose the best response. We report average@3.

\paragraph{RewardBench} RewardBench \citep{lambert2025rewardbench} is a benchmark designed to evaluate the capabilities and safety of reward models. It contains 2,984 cases, divided into four domains: Chat, Chat-Hard, Safety, and Reasoning. Each case contains \textbf{a prompt, one chosen response, and one rejected response}, generated by multiple LLMs.

We split 100 cases from each of the four domains as the evolving set and use the rest as the test set. We provide judges with the prompt and the shuffled two responses, optionally accompanied by a skill, and ask them to choose the best response. We report average@3.

\paragraph{RM-Bench} RM-Bench \citep{liu2024rmbench} is a benchmark designed to evaluate RMs' sensitivity to subtle content differences and resistance to style biases. It contains 1,327 cases, divided into four domains: Chat, Code, Math, and Safety. Each case contains \textbf{a prompt, three chosen responses, and three rejected responses}, generated by multiple LLMs with style control. Each prompt has responses in three styles: concise, detailed, and detailed with Markdown formatting. Combining chosen and rejected responses yields nine response pairs per prompt with different levels of difficulty and we report the average over the $3 \times 3$ chosen--rejected grids.

We split 50 cases from the Chat subset, due to its small size of 129 cases, and 100 cases from each of the other three subsets as the evolving set, using the rest as the test set. For the evolving set, we randomly select one chosen response and one rejected response for each case for skill generation. For the test set, we evaluate each judge on every response-pair combination by providing the prompt and the shuffled two responses, optionally accompanied by a skill, and asking the judge to choose the better response.

\paragraph{JudgeBench} JudgeBench \citep{tan2024judgebench} is a benchmark for evaluating LLM-based judges on challenging response pairs spanning knowledge, reasoning, math, and coding. It contains 350 cases generated by GPT-4o \citep{hurst2024gpto} and 270 cases generated by Claude-3.5-Sonnet. The cases come from varied sources, and each subset is small, so we treat the whole benchmark as one mixed domain to test \texttt{Eval-Skill} in heterogeneous settings. Each case contains \textbf{a prompt, one chosen response, and one rejected response}, generated by multiple LLMs.

We merge and shuffle the cases from both source models, split 100 cases as the evolving set, and use the rest as the test set, while ensuring that the same prompt does not appear in both sets. We provide judges with the prompt and the shuffled two responses, optionally accompanied by a skill, and ask them to choose the better response. We report average@3.

\paragraph{HealthBench} HealthBench \citep{arora2025healthbench} is a benchmark designed to evaluate LLMs on healthcare-related tasks. It contains 5,000 cases, divided into seven themes: Global health, Responding under uncertainty, Expertise-tailored communication, Context seeking, Emergency referrals, Health data tasks, and Response depth. This benchmark is not a reward modeling benchmark, and each case contains \textbf{a single-turn or multi-turn prompt, one human-verified rubric, and various tags}. Each rubric contains multiple criteria, each with a positive or negative score; the final case score is divided by the maximum possible score, so the maximum score is 1.

We adapt the benchmark for reward modeling evaluation by iteratively generating candidate responses with Qwen3-8B for each query and scoring each response with a strong grader using the provided HealthBench rubric. If the current responses differ by less than 0.1, we continue generating and scoring new responses until we obtain at least two responses for the same query whose scores differ by at least 0.1. In this way, we obtain one chosen response with a higher score and one rejected response with a lower score for each retained case.

For the resulting adapted dataset, we split 100 cases from each of the seven themes as the evolving set and use the rest as the test set. We provide judges with the prompt and the shuffled two responses, optionally accompanied by a skill, and ask them to choose the better response. We do not provide judges with the rubrics, since the criteria should be inferred by the judge or hinted by the skill. We report average@3.

\subsection{Baseline Details}
\label{aapx}

For simplicity, we use a unified prompt for both general-purpose and post-trained models, instead of adopting the prompts from the corresponding original papers. Unless otherwise specified, all results are obtained by running the corresponding methods/models ourselves.

\subsection{Model Implementation}
\label{aapx:model_implementation}
In our experiments, we use the following open-weight models:

\begin{itemize}
    \item \textbf{General Models}: Qwen3-4B \citep{yang2025qwen}, Qwen3-8B, Qwen3.6-35B-A3B \citep{qwen36_35b_a3b}, Qwen2.5-7B-Instruct \citep{yang2024qwen}, Llama-3.1-8B \citep{llama31}, Gemma-4-E4B-IT \citep{gemma4}, DeepSeek-V4-Flash \citep{deepseek-ai2026deepseekv}, DeepSeek-V4-Pro.
    \item \textbf{Post-Trained RMs}: RRM-7B \citep{guo2025reward}, RRM-32B, RM-R1-Qwen2.5-Instruct-7B \citep{chen2025rmr}, RM-R1-DeepSeek-Distilled-Qwen-7B, Rubric-RM-Judge-4B \cite{liu2025openrubrics}, Rubric-RM-Judge-8B\footnote{For Rubric-RM models, we use the v2 version from \url{https://huggingface.co/collections/OpenRubrics/rubricrm-v2}.}, and Rubric-ARM-Judge-8B.
    \item \textbf{Post-Trained Rubric Generators}: Rubric-RM-Rubric-4B and Rubric-RM-Rubric-8B.
\end{itemize}
    
Except for the DeepSeek models, which are called via the official API, all models are deployed locally via vLLM and downloaded from \texttt{Hugging Face}. We use non-thinking mode for evaluation unless otherwise specified, and use DeepSeek-V4-Flash with thinking enabled for skill management. We set the temperature to 0.7 for all models. Because we observe occasional repetition in some cases, we set \texttt{max\_completion\_tokens} to 10,000.

Depending on model size, we use NVIDIA RTX A6000, NVIDIA RTX PRO 6000 Blackwell, and Ascend 910B for our experiments. With the \texttt{Eval-Skill} pipeline, generating a skill typically takes 1 to 2 hours using 1 to 2 GPUs, due to the small evolving set size.

\section{Experimental Details}
\label{apx:experimental_details}

\subsection{Main Experiment}
\label{aapx:main_experiment}
Due to space constraints, Table~\ref{tab:main_results} in \S~\ref{ssec:main_experiment_results} reports the main results on RewardBench~2 and RewardBench, while Table~\ref{tab:main_rmbench} provides the RM-Bench results. These results show that \texttt{Eval-Skill} also brings substantial improvements on all of the benchmarks.

\begin{table*}[!t]
\centering
\footnotesize
\begin{tblr}{
width=\linewidth,
colspec={X[1.8,l] X[1.5,l] *{5}{X[0.72,c]}},
colsep=2pt,
rowsep=1pt,
rows={valign=m},
column{2}={leftsep=4pt},
row{1,2}={font=\bfseries},
cell{1}{1}={r=2}{c,m},
cell{1}{2}={r=2}{c,m},
cell{1}{3}={c=5}{c},
hline{1,Z}={1pt,solid},
hline{2}={3-7}{0.5pt,solid,endpos,lr},
hline{3,4,10,11,20,21,25,26}={0.6pt,solid},
}
{Judge~Model} & {Rubric~Model} & RM-Bench & & & & \\
& & Chat & Code & Math & Safety & Avg. \\
\SetCell[c=7]{l} \textbf{Judge Backbone: Qwen3-4B and Variants} & & & & & & \\
Qwen3-4B & w/o & \textbf{67.93} & 54.17 & 70.14 & 85.24 & 69.37 \\
Qwen3-4B & Rubric-RM-4B & 63.71 & 55.47 & 66.23 & 69.01 & 63.61 \\
Qwen3-4B & DSV4 & 62.17 & \uline{57.81} & 71.38 & 87.68 & 69.76 \\
Rubric-RM-4B & Rubric-RM-4B & 48.10 & 51.82 & 48.59 & 63.83 & 53.09 \\
\SetRow{bg=gray!15} Qwen3-4B + Naive~Skill & w/o & 65.68 & 56.86 & \uline{72.57} & \textbf{88.53} & \uline{70.91} \\
\SetRow{bg=gray!15} Qwen3-4B + Eval-Skill & w/o & \uline{67.65} & \textbf{62.84} & \textbf{79.67} & \uline{87.88} & \textbf{74.51} \\
\SetCell[c=7]{l} \textbf{Judge Backbone: Qwen3-8B and Variants} & & & & & & \\
Qwen3-8B & w/o & 66.95 & 59.98 & 70.40 & 90.22 & 71.89 \\
Qwen3-8B & Rubric-RM-8B & 63.43 & 56.94 & 67.91 & 75.53 & 65.95 \\
Qwen3-8B & Rubric-ARM-8B & 63.63 & 57.29 & 67.68 & 77.48 & 66.52 \\
Qwen3-8B & DSV4 & 62.73 & 58.33 & 71.72 & 89.41 & 70.55 \\
Rubric-RM-8B & Rubric-RM-8B & 64.70 & 60.50 & 69.85 & 54.19 & 62.31 \\
Rubric-ARM-8B & Rubric-ARM-8B & 71.17 & \textbf{63.98} & 70.73 & 68.33 & 69.06 \\
\SetRow{bg=gray!15} Qwen3-8B + Naive~Skill & w/o & 70.89 & 57.81 & 72.05 & 90.52 & 72.82 \\
\SetRow{bg=gray!15} Qwen3-8B + Eval-Skill & w/o & \uline{74.40} & 59.98 & \textbf{82.36} & \textbf{92.60} & \uline{77.34} \\
\SetRow{bg=gray!15} Rubric-ARM-8B + Eval-Skill & w/o & \textbf{77.78} & \uline{62.41} & \uline{79.07} & \uline{91.10} & \textbf{77.59} \\
\SetCell[c=7]{l} \textbf{Judge Backbone: DeepSeek-V4-Flash} & & & & & & \\
DSV4 & w/o & 74.96 & 72.05 & 78.30 & 92.60 & 79.48 \\
DSV4 & DSV4 & 76.79 & 76.48 & 79.69 & 90.00 & 80.74 \\
\SetRow{bg=gray!15} DSV4 + Naive Skill & w/o & \uline{78.06} & \textbf{83.33} & \uline{91.66} & \textbf{96.68} & \uline{87.43} \\
\SetRow{bg=gray!15} DSV4 + Eval-Skill & w/o & \textbf{79.61} & \uline{79.95} & \textbf{95.60} & \uline{96.12} & \textbf{87.82} \\
\SetCell[c=7]{l} \textbf{Judge Backbone: Other Post-Trained RMs} & & & & & & \\
RM-R1-Qwen2.5-7B & w/o & 61.74 & 53.30 & 58.64 & 80.51 & 63.55 \\
RM-R1-DS-Qwen-7B & w/o & 67.09 & 54.86 & 81.59 & 80.03 & 70.89 \\
RRM-7B & w/o & 62.31 & 53.73 & 75.24 & 80.06 & 67.84 \\
RRM-32B & w/o & 71.03 & 71.44 & 88.19 & 91.82 & 80.62 \\
\end{tblr}

\caption{Main experiment results on RM-Bench. Within each judge group, the highest score is bolded and the second-highest score is underlined.}
\label{tab:main_rmbench}
\end{table*}

\subsection{Sequential Scaling}
\label{aapx:sequential_scaling}
For sequential scaling, we iteratively refine the skill using rollouts guided by the skill from the previous round. To isolate the effect of iteration, we adopt the one-stage full-skill setting and remove the first-round supervision described in \S~\ref{ssec:ablation}. This yields three effective stages: local skill iteration before merging (pre-merge), hierarchical merging, and global skill iteration after merging (post-merge). We vary the number of generation/refinement iterations, $k_\text{iter}$, from 0 to 6.

In pre-merge refinement, each additional iteration either generates local skills in the first round or refines the local skills before they are hierarchically merged. Here, $k_\text{iter}=0$ corresponds to using no skill, $k_\text{iter}=1$ corresponds to skill generation without further refinement, and $k_\text{iter}>1$ corresponds to $k_\text{iter}-1$ rounds of refinement.

In post-merge iteration, each additional iteration refines an already merged global skill. Inspired by Trace2Skill \citep{ni2026traceskill}, for each round, we first generate patch proposals for each batch, hierarchically merge them, and then apply the merged patch to the previously obtained skill.

Figure~\ref{fig:sequential_scaling} shows that neither Best-in-$k_\text{iter}$ nor Last-in-$k_\text{iter}$ yields a stable monotonic gain on the test set.

\subsection{Parallel Scaling}
\label{aapx:parallel_scaling}

For simplicity, we use the same setting as \texttt{Eval-Skill} and scale three hyperparameters: $k_\text{gen}$, $k_\text{sel}$, and $k_\text{smp}$. In the main experiment, one of the $k_\text{sel}$ branches is obtained by crossing over the top $k_\text{sel}-1$ branches. To isolate the effect of parallel scaling, we remove this crossover operation here and simply select the top $k_\text{sel}$ branches. Note that $k_\text{sel}$ cannot exceed $k_\text{gen}$.

For $k_\text{gen}$, we select the best workflow-only skill from the $k_\text{gen}$ diverse newly generated skills. For $k_\text{sel}$, we select the best full skill from the $k_\text{sel}$ branches. For $k_\text{smp}$, we select the best full skill from the $k_\text{smp}$ samples generated from the same branch. Here, ``best'' denotes the skill with the highest accuracy on the full evolving set $\mathcal{S}*\text{evo}$, and we report its accuracy on the test set $\mathcal{S}*\text{test}$. The branch that performs best on $\mathcal{S}*\text{evo}$ does not necessarily perform best on $\mathcal{S}*\text{test}$, although we find the two accuracies to be strongly correlated.


\subsection{Performance on Other Backbones}
\label{aapx:performance_on_other_backbones}
In addition to Qwen3-4B, Qwen3-8B, and DeepSeek-V4-Flash, which we use in our main experiments, we test \texttt{Eval-Skill} on several other backbone models: (1) Llama-3.1-8B-Instruct, Qwen3.6-35B-A3B, and Gemma-4-E4B-IT, which come from model families other than Qwen3 and DeepSeek; (2) Qwen3-8B with thinking enabled; and (3) RRM-7B, another post-trained model. The results are shown in Table~\ref{tab:other_backbones1} and \ref{tab:other_backbones2}. Foe Eval-Skill, we show the accuracy of the best-performing workflow-only skill from $S_\text{W}$, the best-performing full skill from $S_\text{Full}$, and the final skill chosen between them, based on the performance on the evolving set $\mathcal{S}_\text{evo}$.

\begin{table*}[t]
\centering
\footnotesize
\begin{tblr}{
width=\linewidth,
colspec={X[3.2,l] *{6}{X[0.78,c]} *{5}{X[0.78,c]}},
colsep=2pt,
rowsep=1pt,
hline{1,Z}={1pt,solid},
hline{2}={2-7,8-12}{0.5pt,solid,endpos,lr},
hline{3,4,8,9,13,14,18,19,23,24}={0.5pt,solid},
row{1}={font=\bfseries},
row{2}={font=\bfseries,belowsep=1pt,valign=m},
column{8}={leftsep=6pt},
cell{1}{1}={r=2}{c,m},
cell{1}{2}={c=6}{c},
cell{1}{8}={c=5}{c},
}
Judge~Model & RewardBench~2 & & & & & & RewardBench & & & & \\
& Fact. & Foc. & Math & PIF & Safe. & Avg. & Chat & Hard & Safe. & Rsn. & Avg. \\
\SetCell[c=12]{l} \textbf{Judge Backbone: Llama-3.1-8B-Instruct} & & & & & & & & & & & \\
Direct & 33.60 & 39.66 & \uline{43.78} & \uline{28.33} & 40.86 & 37.25 & \textbf{91.86} & \uline{51.03} & 44.22 & \textbf{83.23} & 67.59 \\
Eval-Skill (workflow-only) & \textbf{41.69} & \textbf{59.49} & 42.57 & \textbf{32.78} & \uline{66.48} & \uline{48.60} & 87.47 & 45.60 & \uline{65.52} & \uline{75.19} & 68.45 \\
Eval-Skill (full) & \uline{40.89} & \uline{42.11} & \textbf{44.58} & 26.11 & \textbf{83.05} & 47.35 & \uline{87.73} & \textbf{53.65} & \textbf{74.06} & 73.76 & \uline{72.30} \\
\SetRow{bg=gray!15} Eval-Skill (final) & \uline{40.89} & \textbf{59.49} & 42.57 & 26.11 & \textbf{83.05} & \textbf{50.42} & \uline{87.73} & \textbf{53.65} & \textbf{74.06} & \uline{75.19} & \textbf{72.66} \\
\SetCell[c=12]{l} \textbf{Judge Backbone: Gemma-4-E4B-IT} & & & & & & & & & & & \\
Direct & 47.73 & 79.16 & 68.67 & \uline{42.22} & 58.48 & 59.25 & \textbf{95.61} & 70.69 & 87.29 & 86.51 & 85.03 \\
Eval-Skill (workflow-only) & \uline{57.78} & \uline{84.14} & \uline{82.33} & 37.78 & \uline{80.95} & \uline{68.60} & \uline{95.22} & \uline{77.34} & \uline{89.48} & \uline{97.42} & 89.87 \\
Eval-Skill (full) & \textbf{60.53} & \textbf{84.73} & \textbf{83.13} & \textbf{62.22} & \textbf{88.48} & \textbf{75.82} & \textbf{95.61} & \textbf{79.40} & \textbf{90.73} & \textbf{97.44} & \textbf{90.80} \\
\SetRow{bg=gray!15} Eval-Skill (final) & \textbf{60.53} & \textbf{84.73} & \textbf{83.13} & \textbf{62.22} & \textbf{88.48} & \textbf{75.82} & \uline{95.22} & \textbf{79.40} & \textbf{90.73} & \textbf{97.44} & \uline{90.70} \\
\SetCell[c=12]{l} \textbf{Judge Backbone: Qwen3.6-35B-A3B} & & & & & & & & & & & \\
Direct & 75.29 & 85.32 & 85.14 & \uline{63.33} & 77.14 & 77.24 & 93.02 & \uline{83.15} & 91.98 & 98.67 & 91.71 \\
Eval-Skill (workflow-only) & \textbf{80.71} & \textbf{89.37} & \textbf{88.76} & \textbf{78.33} & \uline{94.10} & \textbf{86.25} & \uline{94.06} & 82.12 & \uline{93.96} & \textbf{99.42} & \uline{92.39} \\
Eval-Skill (full) & \uline{77.24} & \uline{86.16} & \uline{85.94} & \textbf{78.33} & \textbf{95.71} & 84.68 & \textbf{95.61} & \textbf{84.55} & \textbf{94.11} & \uline{99.32} & \textbf{93.40} \\
\SetRow{bg=gray!15} Eval-Skill (final) & \textbf{80.71} & \uline{86.16} & \uline{85.94} & \textbf{78.33} & \textbf{95.71} & \uline{85.37} & \textbf{95.61} & \textbf{84.55} & \textbf{94.11} & \uline{99.32} & \textbf{93.40} \\
\SetCell[c=12]{l} \textbf{Judge Backbone: Qwen3-8B-think} & & & & & & & & & & & \\
Direct & 57.60 & 82.87 & \textbf{85.14} & 46.67 & 65.62 & 67.58 & \uline{93.80} & \uline{81.18} & 87.55 & 97.94 & 90.12 \\
Eval-Skill (workflow-only) & \textbf{69.07} & \uline{84.73} & 82.33 & \textbf{49.44} & \uline{93.05} & 75.72 & \textbf{95.61} & \textbf{83.80} & \uline{87.92} & \textbf{98.42} & \uline{91.44} \\
Eval-Skill (full) & \uline{67.38} & \textbf{87.00} & \uline{84.74} & \uline{47.22} & \textbf{94.86} & \uline{76.24} & \uline{93.80} & 78.65 & \textbf{91.20} & \uline{98.35} & 90.50 \\
\SetRow{bg=gray!15} Eval-Skill (final) & \textbf{69.07} & \textbf{87.00} & \uline{84.74} & \uline{47.22} & \textbf{94.86} & \textbf{76.58} & \uline{93.80} & \textbf{83.80} & \textbf{91.20} & \uline{98.35} & \textbf{91.79} \\
\SetCell[c=12]{l} \textbf{Judge Backbone: RRM-7B} & & & & & & & & & & & \\
Direct & \uline{36.00} & \uline{64.64} & 56.63 & 20.56 & 43.43 & 44.25 & \uline{84.63} & 67.60 & 79.38 & 88.17 & 79.95 \\
Eval-Skill (workflow-only) & \textbf{41.51} & \textbf{68.86} & \textbf{59.84} & \textbf{32.22} & \textbf{76.29} & \textbf{55.74} & \textbf{85.92} & \textbf{69.38} & \uline{82.60} & \textbf{88.92} & \uline{81.71} \\
Eval-Skill (full) & 35.20 & 61.77 & \uline{57.83} & \uline{23.33} & \uline{68.10} & 49.25 & 83.20 & \uline{67.70} & \textbf{83.18} & \uline{88.67} & 80.69 \\
\SetRow{bg=gray!15} Eval-Skill (final) & \textbf{41.51} & \textbf{68.86} & \textbf{59.84} & \textbf{32.22} & \uline{68.10} & \uline{54.11} & \textbf{85.92} & \textbf{69.38} & \textbf{83.18} & \uline{88.67} & \textbf{81.79} \\
\end{tblr}
\caption{Performance of Eval-Skill across additional judge backbones on RewardBench~2 and RewardBench. Within each judge group, the highest score is bolded and the second-highest score is underlined.}
\label{tab:other_backbones1}
\end{table*}

\begin{table*}[t]
\centering
\footnotesize
\begin{tblr}{
width=\linewidth,
colspec={X[2.7,l] *{5}{X[0.9,c]}},
colsep=2pt,
rowsep=1pt,
hline{1,Z}={1pt,solid},
hline{2}={2-6}{0.6pt,solid},
hline{3,4,8,9,13,14,18,19,23,24}={0.5pt,solid},
row{1}={font=\bfseries},
row{2}={font=\bfseries,belowsep=1pt,valign=m},
cell{1}{1}={r=2}{c,m},
cell{1}{2}={c=5}{c},
}
Judge~Model & RM-Bench & & & & \\
& Chat & Code & Math & Safe. & Avg. \\
\SetCell[c=6]{l} \textbf{Judge Backbone: Llama-3.1-8B-Instruct} & & & & & \\
Direct & 52.74 & \textbf{50.43} & \textbf{59.52} & 38.09 & 50.20 \\
Eval-Skill (workflow-only) & \uline{58.37} & 47.57 & \uline{59.44} & \uline{67.64} & 58.26 \\
Eval-Skill (full) & \textbf{63.01} & \uline{49.83} & 57.24 & \textbf{69.99} & \textbf{60.02} \\
\SetRow{bg=gray!15} Eval-Skill (final) & \textbf{63.01} & 47.57 & \uline{59.44} & \textbf{69.99} & \uline{60.00} \\
\SetCell[c=6]{l} \textbf{Judge Backbone: Gemma-4-E4B-IT} & & & & & \\
Direct & 64.84 & 69.79 & 78.27 & 91.53 & 76.11 \\
Eval-Skill (workflow-only) & \textbf{76.79} & \textbf{79.86} & \uline{88.58} & \uline{93.74} & 84.74 \\
Eval-Skill (full) & \uline{74.96} & \uline{79.43} & \textbf{90.65} & \textbf{94.20} & \uline{84.81} \\
\SetRow{bg=gray!15} Eval-Skill (final) & \textbf{76.79} & \textbf{79.86} & \textbf{90.65} & \textbf{94.20} & \textbf{85.38} \\
\SetCell[c=6]{l} \textbf{Judge Backbone: Qwen3.6-35B-A3B} & & & & & \\
Direct & 77.78 & 80.90 & 89.74 & 94.17 & 85.65 \\
Eval-Skill (workflow-only) & \textbf{84.11} & \textbf{87.50} & \uline{96.17} & \uline{94.69} & \textbf{90.62} \\
Eval-Skill (full) & \uline{82.28} & \uline{85.59} & \textbf{96.37} & \textbf{95.73} & \uline{89.99} \\
\SetRow{bg=gray!15} Eval-Skill (final) & \uline{82.28} & \uline{85.59} & \textbf{96.37} & \textbf{95.73} & \uline{89.99} \\
\SetCell[c=6]{l} \textbf{Judge Backbone: Qwen3-8B-think} & & & & & \\
Direct & 66.53 & 59.38 & 70.71 & 90.03 & 71.66 \\
Eval-Skill (workflow-only) & \textbf{77.64} & \uline{75.17} & \uline{94.82} & \uline{91.53} & \uline{84.79} \\
Eval-Skill (full) & \uline{77.22} & \textbf{76.22} & \textbf{94.95} & \textbf{91.82} & \textbf{85.05} \\
\SetRow{bg=gray!15} Eval-Skill (final) & \uline{77.22} & \textbf{76.22} & \textbf{94.95} & \textbf{91.82} & \textbf{85.05} \\
\SetCell[c=6]{l} \textbf{Judge Backbone: RRM-7B} & & & & & \\
Direct & \uline{62.31} & 53.73 & 75.24 & 80.06 & 67.84 \\
Eval-Skill (workflow-only) & \textbf{62.45} & \uline{54.34} & \uline{78.43} & \textbf{84.62} & \uline{69.96} \\
Eval-Skill (full) & 58.79 & \textbf{56.68} & \textbf{80.78} & \uline{83.87} & \textbf{70.03} \\
\SetRow{bg=gray!15} Eval-Skill (final) & 58.79 & \textbf{56.68} & \textbf{80.78} & \uline{83.87} & \textbf{70.03} \\
\end{tblr}
\caption{Performance of Eval-Skill across additional judge backbones on RM-Bench. Within each judge group, the highest score is bolded and the second-highest score is underlined.}
\label{tab:other_backbones2}
\end{table*}

\paragraph{Backbone Models from Other Families} We select three backbone models of similar size but different model families and release times for analysis. \texttt{Eval-Skill} brings large gains on both Qwen3-8B and the more recent Qwen3.6-35B-A3B (April~2026) and Gemma-4-E4B-IT (May~2026), but the improvement on Llama-3.1-8B-Instruct is less prominent. Through rollout analysis, we find that Llama-3.1-8B-Instruct, a model released in July~2024, does not follow the workflow specified in the skill as reliably, indicating that the backbone's instruction-following ability contributes to the performance of skill-based methods.

\paragraph{Thinking Mode} The comparison between the non-thinking and thinking modes of Qwen3-8B indicates that, although the gain is less prominent, a skill can still significantly improve judge performance by clarifying task priorities and steering evaluation.

\paragraph{Post-Trained Model} In addition to Rubric-ARM-8B-Judge, which is used in our main experiment, we test \texttt{Eval-Skill} on another post-trained model: RRM-7B. In contrast to other backbones, the workflow-only version of \texttt{Eval-Skill} almost consistently outperforms the final full-skill version. We attribute this to the fact that RRM-7B has been post-trained to produce relatively rigid paragraph-style reasoning (e.g., it almost always starts with ``Okay, so I need to evaluate \ldots''), which adapts poorly to the skill. By contrast, Rubric-ARM-8B-Judge is designed to evaluate based on a set of rubrics and can adapt to our skill more easily.

\paragraph{Takeaway} Our skill-based method applies across diverse model types, but stronger instruction-following ability helps models use skills more reliably.

\subsection{Transferability across Backbones}
\label{aapx:transferability_across_backbones}
Table~\ref{tab:trans_model_all} evaluates whether a skill evolved with one backbone can be reused by another backbone. The results show that skills are generally not tied to a single generator: skills generated from rollouts of one model can bring similar gains to models with different capabilities, and they significantly outperform the no-skill baseline.

\begin{table*}[t]
\centering
\footnotesize
\begin{tblr}{
width=\linewidth,
colspec={X[2.1,l] *{5}{X[1,c]} *{5}{X[1,c]}},
colsep=2pt,
rowsep=1pt,
hline{1}={1pt,solid},
hline{2}={2-6,7-11}{0.6pt,solid,endpos,lr},
hline{3,Z}={0.6pt,solid},
row{1-2}={font=\bfseries},
row{2}={valign=m},
column{1}={font=\bfseries},
column{7}={leftsep=6pt},
cell{1}{1}={r=2}{c,m},
cell{1}{2}={c=5}{c,m},
cell{1}{7}={c=5}{c,m},
}
\makecell[l]{Judge\textbackslash{}Skill from} & RewardBench 2 & & & & & RewardBench & & & & \\
& Direct & Q3-4B & Q3-8B & Q3.5-35B & DS-V4-Flash & Direct & Q3-4B & Q3-8B & Q3.5-35B & DS-V4-Flash \\
Q3-4B & 55.80 & \uline{67.55} & \textbf{68.24} & 61.04 & 63.07 & 80.38 & \textbf{85.52} & \uline{84.86} & 81.80 & 78.71 \\
Q3-8B & 57.04 & \textbf{72.17} & \uline{70.48} & 62.25 & 69.46 & 82.39 & \textbf{87.11} & \uline{86.97} & 83.64 & 83.11 \\
Q3.5-35B & 77.24 & \uline{86.64} & 86.07 & 85.37 & \textbf{87.09} & 91.71 & \uline{93.41} & 93.14 & 93.40 & \textbf{94.22} \\
DS-V4-Flash & 68.27 & 84.51 & \textbf{87.06} & 81.59 & \uline{86.78} & 89.66 & \uline{93.35} & 92.91 & 92.47 & \textbf{93.75} \\
\end{tblr}
\caption{Transferability of skills generated by Eval-Skill across different backbone reward models on RewardBench~2 and RewardBench. Within each row and benchmark block, the highest score is bolded and the second-highest score is underlined.}
\label{tab:trans_model_all}
\end{table*}


\subsection{Experiment on HealthBench}
\label{aapx:experiment_on_healthbench}

\begin{table*}[!t]
\centering
\footnotesize
\begin{tblr}{
width=\linewidth,
colspec={X[2.0,l] *{8}{Q[c,wd=0.1\linewidth]}},
colsep=0.8pt,
rowsep=1pt,
rows={valign=m},
hline{1,Z}={1pt,solid},
hline{2}={2-9}{0.5pt,solid,endpos,lr},
hline{5,8}={0.6pt},
hline{3}={0.6pt,solid},
row{1-2}={font=\bfseries},
column{1}={font=\bfseries},
cell{1}{1}={r=2}{c,m},
cell{1}{2}={c=8}{c,m},
}
Methods & HealthBench & & & & & & & \\
& Comm. & \makecell{Complex\\Responses} & \makecell{Context\\Seeking} & \makecell{Emergency\\Referrals} & \makecell{Global\\Health} & \makecell{Health Data\\Tasks} & Hedging & Avg. \\
Qwen3-8B & 51.60 & 53.75 & 55.38 & 51.31 & 57.17 & 51.30 & 54.02 & 53.50 \\
Qwen3-8B + Eval-Skill & 52.84 & 54.97 & 53.16 & 52.53 & 56.82 & 53.37 & 56.41 & 54.30 \\
Qwen3.6-35B-A3B & 60.99 & \textbf{59.51} & \textbf{57.26} & 56.37 & 58.72 & 55.44 & 54.87 & 57.59 \\
Qwen3.6-35B-A3B (Rubric-Based) & 60.46 & \textbf{59.51} & 52.82 & 57.42 & 55.79 & 55.09 & 59.32 & 57.20 \\
Qwen3.6-35B-A3B + Skill from Qwen3-8B & 58.16 & 56.20 & 56.58 & \textbf{61.08} & 57.51 & \textbf{59.24} & \textbf{62.91} & \textbf{58.81} \\
DeepSeek-V4-Flash & 53.55 & 51.31 & 49.57 & 48.52 & 58.55 & 52.68 & 56.75 & 52.99 \\
DeepSeek-V4-Flash + Skill from Qwen3-8B & \textbf{61.35} & 56.37 & 55.38 & \textbf{61.08} & \textbf{60.62} & 57.17 & 56.58 & 58.36 \\
\end{tblr}

\caption{Performance on HealthBench across its seven themes. To simplify the setup and leverage cross-backbone skill transfer (\S~\ref{ssec:other_backbones}), we use the skill generated from Qwen3-8B for all models. The rubric model uses self-generated rubrics rather than the ground-truth HealthBench rubrics, which are not provided to any judge at evaluation time.}
\label{tab:app_hb}
\end{table*}

Table~\ref{tab:app_hb} shows that \texttt{Eval-Skill} does not yield consistent gains on HealthBench. For Qwen3-8B and Qwen3.6-35B-A3B, the average gain is small, and performance decreases on several themes. Still, the skill transferred from Qwen3-8B improves the performance of DeepSeek-V4-Flash, although to a smaller degree than in the main experiments. These results suggest that HealthBench, combined with the ``choose candidates from the same actor model'' setting, remains difficult for skill-guided reward modeling.


\subsection{Comparison with Skill-RM}
\label{aapx:comparison_with_skill_rm}

We compare Eval-Skill with Skill-RM~\citep{chen2026skillrm} on the same benchmarks used by Skill-RM: RewardBench~2, RM-Bench, and JudgeBench. As shown in Table~\ref{tab:skill_rm_1}~and~\ref{tab:skill_rm_2}, Eval-Skill generally outperforms Skill-RM in both accuracy and accuracy improvement over the vanilla baseline ($\Delta_{\text{acc}}$). However, the benchmark protocols differ in several aspects:

\begin{itemize}
\item \textbf{RewardBench~2:} As in the main experiments, we exclude the Tie subset and retain the other four subsets: Factuality, Focus, Precise IF, and Safety. In contrast, Skill-RM includes the Tie subset. We report average@3, whereas Skill-RM reports best-of-four.

\item \textbf{RM-Bench:} We require the judge to always choose the better answer from each pair, while Skill-RM allows ties. Both Eval-Skill and Skill-RM report the average over the $3 \times 3$ chosen--rejected grids.

\item \textbf{JudgeBench:} We report average@3, whereas Skill-RM reports best-of-two, with ties counted as incorrect. Both Eval-Skill and Skill-RM use the GPT and Claude subsets.

\end{itemize}

We also note that, for each domain in each subset, we reserve a portion of the cases as the evolving set for skill synthesis and use the remaining cases for testing; see Appendix~\ref{aapx:benchmarks}. In contrast, Skill-RM uses handcrafted skills and evaluates on the full datasets. For the weighted comparison, however, we use the original pre-split weights to ensure a fair comparison.

\begin{table*}[!t]
\centering
\footnotesize
\begin{tblr}{
  width=\linewidth,
  colspec={X[4,l] *{9}{X[c]}},
  cells={c},
  column{1}={l},
  rows={valign=m},
  row{1-2}={font=\bfseries},
  hline{1,Z}={1pt,solid},
  hline{2}={0.5pt,solid,endpos,lr},
  hline{3}={0.5pt,solid},
  hline{6}={0.5pt,dashed},
}
\SetCell[r=2]{c}\textbf{Methods}
  & \SetCell[c=9]{c}\textbf{RewardBench 2}
  & & & & & & & & \\
  & \textbf{Fac.}
  & \textbf{Focus}
  & \textbf{Math}
  & \makecell{\textbf{Precise}\\\textbf{IF}}
  & \textbf{Safety}
  & \textbf{Tie}
  & \textbf{Avg.}
  & \textbf{Weighted}
  & \makecell{\textbf{$\Delta{}_\text{acc}$}} \\
Qwen3.5-27B*
  & -
  & -
  & -
  & -
  & -
  & -
  & -
  & 81.10
  & - \\
Qwen3.5-27B+Skill-RM*
  & -
  & -
  & -
  & -
  & -
  & -
  & -
  & 85.00
  & +3.90 \\
  \quad + sample-spec.
  & -
  & -
  & -
  & -
  & -
  & -
  & -
  & 86.00
  & +4.90 \\
Qwen3.5-27B
  & 79.91
  & 87.00
  & 85.94
  & 73.89
  & 82.48
  & $\times$
  & 81.84
  & 82.64
  & - \\
Qwen3.5-27B+LESS-Eval
  & 82.40
  & 93.76
  & 91.97
  & 75.56
  & 98.19
  & $\times$
  & 88.38
  & \textbf{89.99}
  & \textbf{+7.36} \\
\end{tblr}
\caption{Performance comparison between Eval-Skill and Skill-RM on RewardBench~2. Results marked with * are reported by the original paper of Skill-RM.}
\label{tab:skill_rm_1}
\end{table*}

\begin{table*}[!t]
\centering
\footnotesize
\begin{tblr}{
  width=\linewidth,
  colspec={X[4,l] *{9}{X[c]}},
  cells={c},
  column{1}={l},
  rows={valign=m},
  row{1-2}={font=\bfseries},
  hline{1,Z}={1pt,solid},
  hline{2}={0.5pt,solid,endpos,lr},
  hline{3}={0.5pt,solid},
  hline{6}={0.5pt,dashed},
}
\SetCell[r=2]{c}\textbf{Methods}
& \SetCell[c=7]{c}\textbf{RM-Bench}
& & & & & &
& \SetCell[c=2]{c}\textbf{JudgeBench}
& \\
  & \textbf{Chat}
  & \textbf{Code}
  & \textbf{Math}
  & \textbf{Safety}
  & \textbf{Avg.}
  & \textbf{Weighted}
  & \makecell{\textbf{$\Delta{}_\text{acc}$}}
  & \textbf{-}
  & \makecell{\textbf{$\Delta{}_\text{acc}$}} \\
Qwen3.5-27B*
  & -
  & -
  & -
  & -
  & -
  & 89.80
  & -
  & 80.80
  & - \\
Qwen3.5-27B+Skill-RM*
  & -
  & -
  & -
  & -
  & -
  & 91.50
  & +1.70
  & 82.10
  & +1.30 \\
  \quad + sample-spec.
  & -
  & -
  & -
  & -
  & -
  & 91.50
  & +1.70
  & 89.70
  & \textbf{+7.60} \\
Qwen3.5-27B
  & 76.93
  & 81.45
  & 93.59
  & 94.87
  & 86.71
  & 90.31
  & -
  & 85.00
  & - \\
Qwen3.5-27B+LESS-Eval
  & 79.89
  & 86.98
  & 97.10
  & 97.17
  & 90.29
  & \textbf{93.71}
  & \textbf{+3.40}
  & \textbf{90.71}
  & +5.71 \\
\end{tblr}
\caption{Performance Comparison between Eval-Skill and Skill-RM on RM-Bench and JudgeBench. Results marked with * are reported by the original paper of Skill-RM.}
\label{tab:skill_rm_2}
\end{table*}

\subsection{Skill Template and Examples}
\label{aapx:skill_examples}
We show the full skill template and several skills generated by \texttt{Eval-Skill} to clarify the skill structure and the effect of \texttt{Eval-Skill}. 

\begin{itemize}
    \item The full skill template is shown in Figure~\ref{fig:skill_template}.
    \item The skill from Qwen3-8B on the Factuality domain of RewardBench~2 is shown in Figure~\ref{fig:skill_q8_factuality}.
    \item The skill snippet from Qwen3-4B on JudgeBench is shown in Figure~\ref{fig:skill_q4_jb}.
\end{itemize}

\FloatBarrier
\clearpage

\begin{figure*}[ptb]
\centering
\begin{tcolorbox}
\textbf{Full Skill Template}
\medskip\hrule\medskip
\begin{Verbatim}[breaklines=true,breakanywhere=true,fontsize=\small,breaksymbolleft={},breaksymbolright={}]
name: [SkillName]
description: [Clear, concise description of what this skill does and when to use it.
1-2 sentences focusing on the core purpose and benefits.]
version: 1.0.0
---
# [Skill Title]
## Overview
[1-2 sentences on the core approach]

## Workflow
1. **Analysis**: [Instruct the RM to conduct its detailed analysis under the `--- Analysis ---` section. DO NOT just copy this hint; specify the exact step-by-step format and output layout the RM should use (e.g., examining each candidate response or criteria one by one, applying this skill's principles and the generated rubric, while following the core intent of the user instruction).]
2. **Final Judgment**: [Instruct the RM to aggregate findings under the `--- Final Judgment ---` section and explicitly select exactly ONE winner (never "None" or "Neither"). DO NOT just copy this hint; provide concrete guidance on how to weigh the criteria and aggregate findings to pick the winner, generating the exact fields `Aggregation Summary:`, `Justification:`, and `Winner:`.]

## Principles
1. **[Short Name]**
**Condition**: [Trigger condition for the principle]
**Principle**: [How to implement the principle in evaluating candidate responses. Must not contradict the forced-choice rule.]
**Example**: [One or more concise example(s) illustrating the condition and the principle. Don't include case IDs or specific options. ]
**Anti-Pattern**: [What to avoid when applying this principle]
2. ...
\end{Verbatim}
\end{tcolorbox}
\caption{Full skill template.}
\label{fig:skill_template}
\end{figure*}

\begin{figure*}[ptb]
\centering
\begin{tcolorbox}
\textbf{Skill Generated for Qwen3-8B on Factuality Domain of RewardBench~2}
\medskip\hrule\medskip

\begin{Verbatim}[breaklines=true,breakanywhere=true,fontsize=\small,breaksymbolleft={},breaksymbolright={}]
name: Factual-Integrity and Adherence Evaluator
description: Evaluates candidate responses by prioritizing verifiable factual accuracy, strict instruction compliance, domain relevance, and appropriate hedging, with logical coherence as a tiebreaker. Ideal for knowledge-based, technical, or instruction-following queries where objective correctness and precise fulfillment of user intent are critical.
version: 1.0.0
---
# Factual-Integrity and Adherence Evaluator
## Overview
A systematic evaluation that first verifies every factual claim, rejects hallucinations, and enforces explicit user constraints. Only after primary criteria are satisfied do we compare logical structure, clarity, and context relevance. The factually sound, compliant, and honest response wins, even if shorter or less polished than alternatives tainted by errors.

## Workflow
1. **Analysis**: Under the `--- Analysis ---` section, evaluate each candidate independently using the following step-by-step criteria in order of priority. For each response, write:
   - **`Response [Letter]:`** Justification: [Factual Integrity & Hallucination Check (verify every claim, flag major/minor errors, fabricated details, implausible technical claims, anachronisms); Safety & Ethical Boundaries (for sensitive topics, note safe refusal vs harmful speculation); Instruction Compliance (check explicit constraints: counts, format, date ranges, “no explanations”, language, literal interpretation – violations heavily penalized); Domain & Context Relevance (tailored to implied audience/field/culture; overly technical or off-topic is weakness); Hedging & Uncertainty (prefer cautious language for unverifiable topics; separate verifiable from unverifiable; flag confident falsehoods); Logical Coherence & Clarity (tie-breaker; note contradictions, fallacies, organization)]
   Example: **Response A:** Justification: Factual: two major hallucinations (fabricated entity, wrong date); Safety: safe refusal on harmful query; Compliance: violated “no explanations” rule; Domain: off-topic for nursing research; Hedging: overconfident on unverifiable claim; Logic: clear but based on false premises. Summary: heavily flawed, cannot win.

2. **Final Judgment**: Under `--- Final Judgment ---`:
   - `Aggregation Summary:` Compare all candidates primarily by factual integrity and safety. Eliminate any with major factual errors or hallucinations. Among remaining, prioritize instruction compliance and domain relevance. Then consider hedging/honesty. If tied, use logical coherence. Highlight which response is most reliable and fulfills user intent.
   - `Justification:` Explain why the chosen response is superior, referencing specific errors (hallucinations, instruction violations) in others. Address why a shorter or less polished response wins if factually sound and compliant.
   - `Winner:` Exactly one letter (e.g., `Winner: Response C`). Never "None" or "Neither". If two responses are equally good, choose the one with best factual integrity then instruction fidelity. The analysis must support a clear decision.

## Principles
1. **Factual Primacy and Hallucination Rejection**
   **Condition**: When a response makes specific factual claims.
\end{Verbatim}
\begin{flushright}
\textcolor{black!55}{\emph{(continue on next page)}}
\end{flushright}
\end{tcolorbox}
\end{figure*}
\begin{figure*}
\begin{tcolorbox}
\textcolor{black!55}{\emph{(continue from previous page)}} \\[0.5cm]
\begin{Verbatim}[breaklines=true,breakanywhere=true,fontsize=\small,breaksymbolleft={},breaksymbolright={}]
   **Principle**: Factual accuracy is the gatekeeper. A response with any hallucinated detail (fabricated dates, entities, events, sources) or verifiable major error is automatically inferior to one without such flaws, regardless of style or length. Minor imprecisions acceptable only if no better response exists. Treat implausible or unsupported specifics as hallucinations. For technical topics, check plausibility.
   **Example**: For "What is the main ingredient in chocolate?", a short "cocoa beans" beats a long essay claiming "cocoa beans and coconut oil" (error). For a query about iPhone 6s Plus, a claim of 10x zoom via software update is a hallucination.
   **Anti-Pattern**: Do not reward elaborate detail or confident tone that masks errors. Do not accept technical specifics at face value.

2. **Strict Instruction Compliance**
   **Condition**: When user gives explicit commands, constraints, or uses strong modifiers.
   **Principle**: Evaluate compliance rigorously. Violations of explicit constraints (count, format, date range, "no explanations", literal interpretation) are heavily penalized. Prefer responses that consistently apply rules (e.g., antonym replacement, verbatim quotes). A single instruction violation can disqualify if a compliant alternative exists.
   **Example**: For "List 20 musical groups from the 2000s in random order", a response with 20 correct groups in random order beats one that includes 2010s groups or lists chronologically. For "write literal opposites of movie titles", a response that changes "The Silence of the Lambs" to "The Loudness of the Lions" beats one that injects a new character.
   **Anti-Pattern**: Do not ignore a constraint violation because the response looks comprehensive or polished. Do not favor a well-structured list if examples violate the rule.

3. **Safety and Ethical Boundaries**
   **Condition**: When query involves sensitive, harmful, or exploitative content.
   **Principle**: Prioritize safe refusal over speculative or fabricated responses. A response that refuses to engage with harmful content is superior to one that invents details. Safety violations (e.g., generating misinformation about a person) are critical errors.
   **Example**: For a query about a person with exploitative content, a response stating "I cannot answer that as it may be harmful" beats one that fabricates a biography.
   **Anti-Pattern**: Do not penalize a safe refusal as evasive.

4. **Domain & Context Relevance**
   **Condition**: When user implies a specific audience, field, or cultural perspective.
   **Principle**: Prefer responses directly relevant to that context. Overly technical details from a different specialty or generic explanations ignoring cultural specifics are weaknesses. Domain alignment is a primary criterion after factual accuracy and instruction compliance.
   **Example**: For "give 10 topics for spinal anesthesia research in nursing", a response focusing on patient education and interprofessional collaboration wins over one focusing on drug dosage comparisons (anesthesiology domain).
   **Anti-Pattern**: Do not favor a response that is "more detailed" if it does not address user's implied role or context.

5. **Temporal and Definitional Precision**
   **Condition**: When query specifies a time period, decade, or asks for a definition/trend.
   **Principle**: Strictly enforce the given time window – any item clearly outside the period is a major anachronism. For definitions, verify standard scientific wording. Inverted trends are major errors.
   **Example**: Query "List similar things in the 1990s." A response includes Roblox (2006) – anachronism; the response that only lists 1990s items wins.
   **Anti-Pattern**: Do not dismiss an anachronism as minor; it indicates unreliability.
\end{Verbatim}
\begin{flushright}
\textcolor{black!55}{\emph{(continue on next page)}}
\end{flushright}
\end{tcolorbox}
\end{figure*}
\begin{figure*}
\begin{tcolorbox}
\textcolor{black!55}{\emph{(continue from previous page)}} \\[0.5cm]
\begin{Verbatim}[breaklines=true,breakanywhere=true,fontsize=\small,breaksymbolleft={},breaksymbolright={}]
6. **Appropriate Hedging for Unverifiable Content**
   **Condition**: When query asks about supernatural, subjective, or legendary topics.
   **Principle**: Favor responses using cautious language ("it is said", "many believe") over those presenting such claims as established facts. However, hedging does not excuse factual errors in verifiable details (geography, history). Separate unverifiable from verifiable claims.
   **Example**: For "Montserrat is a center of teluric energy", a response with "some consider" is better than without hedging. But including Mallorca when asked for Catalonia is a factual error even if hedged.
   **Anti-Pattern**: Do not assume all hedging is equal; verify that response does not contain implicit factual assertions.

7. **Logical Coherence and Clarity as Tie‑Breakers**
   **Condition**: After factual accuracy, safety, compliance, domain relevance, and hedging are satisfied, multiple responses remain equally correct.
   **Principle**: Use logical structure, clarity, absence of contradictions, and conciseness as tie‑breakers. Prefer the response with clear thesis and well-organized reasoning. Among factually equal candidates, a concise correct answer beats a verbose correct one.
   **Example**: Two factually correct responses both list the same data; the one with clear headings and logical flow wins over a jumbled paragraph.
   **Anti-Pattern**: Do not use coherence to override factual errors.

8. **Forced-Choice Enforcement**
   **Condition**: Always at final judgment.
   **Principle**: RM must explicitly pick exactly one candidate as winner. Never output "None" or "Neither". If two responses are equally good, choose the one that best follows the first principle (factual integrity) and then instruction fidelity. The analysis must support a clear decision.
   **Example**: In a four-option comparison, if three contain major factual errors and one is factually sound but less detailed, the accurate one wins.
   **Anti-Pattern**: Do not leave judgment ambiguous.
\end{Verbatim}
\end{tcolorbox}
\caption{Generated skill for Qwen3-8B on the Factuality domain of RewardBench~2.}
\label{fig:skill_q8_factuality}
\end{figure*}

\begin{figure*}[ptb]
\centering
\begin{tcolorbox}
\textbf{Skill Snippet Generated for Qwen3-4B on JudgeBench Mixed}
\medskip\hrule\medskip
...

\begin{Verbatim}[breaklines=true,breakanywhere=true,fontsize=\small,breaksymbolleft={},breaksymbolright={}]
## Workflow
1. **Analysis**: Under the `--- Analysis ---` section, for each candidate response, perform the evaluation in the following strict order. Record findings in a concise per-candidate summary.
   - **Mandatory Requirements Compliance**: Identify all explicit format, output structure, and process instructions from the user query (e.g., "use a box", "repeat exactly five times", "think step by step"). Assess whether the response exactly complies. Classify any deviation as:
     - **Critical** - missing required output, wrong count by more than 1, completely wrong structure, ambiguous output, omission of required reasoning when explicitly demanded, or violation of a format that is the primary deliverable (e.g., "output exactly five `A`s").
     - **Minor** - extra punctuation, slight extra descriptive text, off-by-one character if the intended answer remains unambiguous, or minor formatting discrepancies (e.g., bold vs. asterisks) when format is not the primary requirement.
   - **Core Correctness (with Internal Consistency & Premise Fidelity)**: Determine the correct answer or expected outcome **before** examining the candidate's answer, using the method best suited to the query:
\end{Verbatim}
\vspace{-1\baselineskip}
\begin{Verbatim}[breaklines=true,breakanywhere=true,fontsize=\small,breaksymbolleft={},breaksymbolright={},formatcom=\color{green!45!black}]
     - *Domain factual* - recall authoritative ground truth.
     - *Algorithmic / code* - manually simulate at least one sample input step-by-step.
     - *Multi-part / constraint-heavy* - verify coverage of all parts and respect for every given constraint.
     - *Math / numeric* - compute independently with appropriate precision.
\end{Verbatim}
\vspace{-1\baselineskip}
\begin{Verbatim}[breaklines=true,breakanywhere=true,fontsize=\small,breaksymbolleft={},breaksymbolright={}]
     Then compare the candidate's final answer against that ground truth. Additionally, check the reasoning for any self-contradictions, misinterpretation of premises, or logical leaps. A candidate whose final answer matches ground truth but whose reasoning contains a contradiction or violates given premises is deemed incorrect (critical error). Document whether the core answer is correct and whether the reasoning is logically sound.
   - **Precision & Domain Alignment**: If the query involves domain-specific definitions, formulas, or standard test design, verify that the candidate uses precise, standard terminology and correct derivations. Favor responses that align with typical educational or professional context. A vague or imprecise answer (e.g., guessing a numeric value without derivation) loses to a precise one when both are otherwise equal.
   - **Contextual Alignment**: If the query specifies a particular perspective, domain, or framing (e.g., "from a behaviorist view", "main purpose of licensing"), evaluate whether the response correctly adopts that specific angle over a general statement.
   - **Constraint & Feasibility** (algorithmic tasks only): Verify that no disallowed external packages are used; check that time/space complexity is feasible given constraints. For other tasks, ensure no unwarranted assumptions.
   - **Justification**: Write a concise summary for each candidate covering: mandatory requirement status (with severity), core correctness verdict (including any internal inconsistency), precision, contextual alignment, and any critical errors.

2. **Final Judgment**: Under the `--- Final Judgment ---` section, aggregate findings using the following ordered hierarchy. **Always output exactly three fields**: `Aggregation Summary:`, `Justification:`, and `Winner:` (e.g., `Winner: A`). Never output “None” or “Neither”.
\end{Verbatim}

...

\end{tcolorbox}
\caption{Generated skill for Qwen3-4B on JudgeBench Mixed. The workflow adaptively requires the RM to determine the correct evaluation procedure based on the query type.}
\label{fig:skill_q4_jb}
\end{figure*}



\end{document}